\documentclass{article}

\usepackage{microtype}
\usepackage{graphicx}
\usepackage{subfigure}
\usepackage{booktabs} 

\usepackage{hyperref}

\usepackage[accepted]{icml2024}

\usepackage{amsmath,amsfonts,bm}

\def\eqref#1{equation~\ref{#1}}

\def\1{\bm{1}}

\def\vmu{{\bm{\mu}}}
\def\vtheta{{\bm{\theta}}}

\def\vmu{{\bm{\mu}}}
\def\vsigma{{\bm{\sigma}}}

\def\vx{{\bm{x}}}
\def\vxhat{{\bm{\hat{x}}}}

\DeclareMathAlphabet{\mathsfit}{\encodingdefault}{\sfdefault}{m}{sl}
\SetMathAlphabet{\mathsfit}{bold}{\encodingdefault}{\sfdefault}{bx}{n}

\usepackage{amsmath}
\usepackage{amssymb}
\usepackage{mathtools}
\usepackage{amsthm}
\usepackage{graphicx,wrapfig,lipsum}

\usepackage[capitalize,noabbrev]{cleveref}

\theoremstyle{plain}

\theoremstyle{definition}

\theoremstyle{remark}

\usepackage[textsize=tiny]{todonotes}

\begin{document}

\twocolumn[
\icmltitle{All-in-one simulation-based inference
}



\icmlsetsymbol{equal}{*}

\begin{icmlauthorlist}
\icmlauthor{Manuel Gloeckler}{tue}
\icmlauthor{Michael Deistler}{tue}
\icmlauthor{Christian Weilbach}{ubc}
\icmlauthor{Frank Wood}{ubc}
\icmlauthor{Jakob H. Macke}{tue,mpi}
\end{icmlauthorlist}

\icmlaffiliation{tue}{Machine Learning in Science, University of Tübingen and Tübingen AI Center, Tübingen, Germany}
\icmlaffiliation{mpi}{Max Planck Institute for Intelligent Systems, Department Empirical Inference, Tübingen, Germany}
\icmlaffiliation{ubc}{Department of Computer Science, University of British
Columbia, Vancouver, Canada}


\icmlcorrespondingauthor{Manuel Gloeckler}{manuel.gloeckler@uni-tuebingen.de}
\icmlcorrespondingauthor{Jakob H. Macke}{jakob.macke@uni-tuebingen.de}

\icmlkeywords{Machine Learning, ICML}

\vskip 0.3in
]



\printAffiliationsAndNotice{}  

\begin{abstract}
Amortized Bayesian inference trains neural networks to solve stochastic inference problems using model simulations, thereby making it possible to rapidly perform Bayesian inference for any newly observed data. However, current simulation-based amortized inference methods are simulation-hungry and inflexible: They require the specification of a fixed parametric prior, simulator, and inference tasks ahead of time. Here, we present a new amortized inference method---the Simformer---which overcomes these limitations. By training a probabilistic diffusion model with transformer architectures, the Simformer outperforms current state-of-the-art amortized inference approaches on benchmark tasks and is substantially more flexible: It can be applied to models with function-valued parameters, it can handle inference scenarios with missing or unstructured data, and it can sample arbitrary conditionals of the joint distribution of parameters and data, including both posterior and likelihood. We showcase the performance and flexibility of the Simformer on simulators from ecology, epidemiology, and neuroscience, and demonstrate that it opens up new possibilities and application domains for amortized Bayesian inference on simulation-based models.
\end{abstract}

\begin{figure}
    \centering
    \includegraphics[width=0.5\textwidth]{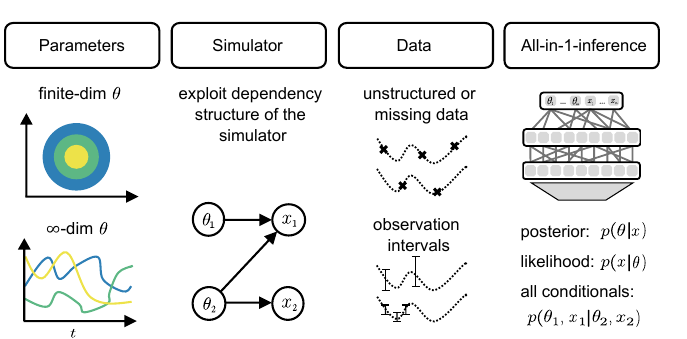}
    \caption{Capabilities of the Simformer: It can perform inference for simulators with a finite number of parameters or function-valued parameters (first column), it can exploit dependency structures of the simulator to improve accuracy (second column), it can perform inference for unstructured or missing data, for observation intervals (third column), and it provides an `all-in-one' inference method that can sample all conditionals of the joint distribution, including posterior and likelihood (fourth column).}
    \label{fig:illustration}
    
\end{figure}

\section{Introduction}

Numerical simulators play an important role across various scientific and engineering domains, offering mechanistic insights into empirically observed phenomena \cite{gonccalves2020training,dax2021gravitational, marlier2021simulationbased}. A fundamental challenge in these simulators is the identification of unobservable parameters based on empirical data, a task addressed by simulation-based inference (SBI) \cite{cranmer2020frontier}, which aims to perform Bayesian inference using samples from a (possibly blackbox) simulator, without requiring access to likelihood evaluations. 
A common approach in SBI is to train a neural network on pairs of parameters and corresponding simulation outputs: After an initial investment in simulations and network training, inference for any observation can then be performed without further simulations. These methods thereby \emph{amortize} the cost of Bayesian inference.

Many methods for amortized SBI have been developed recently \citep{papamakarios2016fast,lueckmann2017flexible,le2017inference,greenberg2019automatic,papamakarios2019sequential,radev2020bayesflow, hermans2020likelihood,gloeckler2022variational,boelts2022flexible,deistler2022truncated,sharrock2022sequential}. While these methods have different strengths and weaknesses, most of them also share limitations.
First, they often rely on structured, tabular data (typically $\vtheta, \vx$ vectors). Yet, real-world datasets are often more messy \cite{shukla2021survey}: Irregularly sampled time series naturally arise in domains like ecology, climate science, and health sciences. Missing values often occur in real-world observations and are not easily handled by existing approaches.  
Second, the inputs of a simulator can correspond to a function of time or space, i.e., $\infty$-dimensional parameters \cite{chen2020sir,ramesh2022gatsbi}. Existing amortized methods typically necessitate discretization, thereby limiting their applicability to a specific, often dense grid and precludes the evaluation of the parameter posterior beyond this grid. 
Third, they require specification of a fixed approximation task: the neural network can either target the likelihood (neural likelihood estimation, NLE, \citet{papamakarios2019sequential}) or the posterior (neural posterior estimation, NPE, \citet{papamakarios2016fast}). In practice, users might want to interactively explore both conditional distributions, investigate posteriors conditioned on subsets of data and parameters, or even explore different prior configurations.
Fourth, while neural-network based SBI approaches are more efficient than classical ABC-methods \cite{lueckmann2021benchmarking}, they are still simulation-hungry. In part, this is because they target blackbox simulators, i.e., they do not require any access to the model's inner workings. However, in practice, one has at least \emph{partial} knowledge (or assumptions) about the structure of the simulator (i.e., its conditional independencies), but common SBI methods cannot exploit such knowledge.
These limitations have prevented the application of SBI in \emph{interactive} applications, in which properties of the task need to be changed on the fly. 


Here, we develop a new method for amortized Bayesian inference---the Simformer---which overcomes these limitations (Fig.~\ref{fig:illustration}), using a combination of transformers and probabilistic diffusion models \cite{peebles2023scalable, hatamizadeh2023diffit}, based on the idea of graphically structure diffusion models proposed by \citet{weilbach2023graphically}. Our method can deal with unstructured and missing data and handles both parametric and nonparametric simulators (i.e., with function-valued $\infty$-dimensional) parameters. In addition, the method returns a single network that can be queried to sample \emph{all} conditionals of the joint distribution (including the posterior, likelihood, and arbitrary parameter conditionals) and can also perform inference if the observations are intervals as opposed to specific values.
We show that our method has higher accuracy than previous SBI methods on benchmark tasks (for a given simulation budget). Moreover, by using attention masks, one can use domain knowledge to adapt the Simformer to the dependency structure of the simulator \cite{weilbach2023graphically} to further improve simulation efficiency. 
Thus, the Simformer provides an `all-in-one` inference method that encapsulates posterior- and likelihood-estimation approaches and expands the space of data, simulators, and tasks for which users can perform simulation-based amortized Bayesian inference.


\label{sec:introduction}

\section{Preliminaries}
\label{sec:background}

\subsection{Problem setting and approach}
We consider a simulator with parameters $\vtheta$ (potentially nonparametric) which stochastically generates samples $\vx$ from its implicit likelihood $p(\vx|\vtheta)$. After having observed data $\vx_o$, we aim to infer the posterior distribution $p(\vtheta | \vx_o)$ of parameters given data, but also retain the flexibility to capture any other conditional of the full joint $p(\vtheta, \vx)$. 
We, therefore, introduce the joint $\vxhat= (\vtheta, \vx)$, that will serve as input for a transformer together with a mask indicating which values are \textit{observed}. The transformer will then use attention mechanisms to compute the corresponding sequence of output scores of equal size. The scores corresponding to \emph{unobserved} variables will then form the basis of a diffusion model representing the distribution over these variables.
We first give background on the main ingredients (transformers and score-based diffusion models) in this section before giving a detailed description in Sec.~\ref{sec:methods}.



\subsection{Transformers and attention mechanisms}
Transformers overcome limitations of feed-forward networks in effectively dealing with sequential inputs. They incorporate an attention mechanism which, for a given sequence of inputs, replaces individual hidden states with a weighted combination of all hidden states \citep{vaswani2017attention}. Given three learnable linear projections of each hidden state ($Q$, $K$, $V$) this is computed as
$$\text{attention}(Q, K, V) = \text{softmax}(Q K)^T / \sqrt{d})V.$$

\subsection{Score-based diffusion models}
\label{subsec:score_sde}

Score-based diffusion models \citep{song2020score,song2020generative} describe the evolution of data through stochastic differential equations (SDEs). 
Common SDEs for score-based diffusion models can be expressed as
$$ d\vxhat_t = f(\vxhat_t, t)dt + g(t) d\bm{w},$$
with $\bm{w}$ being a standard Wiener process, and $f$ and $g$ representing the drift and diffusion coefficients, respectively. The solution to this SDE defines a diffusion process that transforms an initial data distribution $p_0(\vxhat_0) = p(\vxhat)$ into a simpler noise distribution $p_T(\vxhat_T) \approx \mathcal{N}(\vxhat_T;\vmu_T, \vsigma_T)$.

Samples from the generative model are then generated by simulating the reverse diffusion process \citep{anderson1982reverse}
$$ d\vxhat_t = \left[f(\vxhat_t, t) - g(t)^2 s(\vxhat_t,t) \right]dt + g(t)d\bm{\tilde{w}},$$
where $\bm{\tilde{w}}$ is a backward-in-time Wiener process. This relies on the knowledge of the score function $s(\vxhat_t,t)=\nabla_{\vxhat_t} \log p_t(\vxhat_t)$ at each step.
The exact marginal score is typically intractable but can be estimated through time-dependent denoising score-matching \cite{hyvarinen2005estimation, song2020score}. Given that the conditional score is known, $p_t(\vxhat_t|\vxhat_0)=\mathcal{N}(\vxhat_t; \mu_t(\vxhat_0), \sigma_t(\vxhat_0))$, the score model $s_\phi(\vxhat_t, t)$ is trained to minimize the loss
$$ \mathcal{L}(\phi) = \mathbb{E}_{t, \vxhat_0, \vxhat_t} \left[ \lambda(t) \left\lVert s_\phi(\vxhat_t, t) - \nabla_{\vxhat_t} \log p_t(\vxhat_t|\vxhat_0)\right\rVert_2^2 \right], $$
where $\lambda$ denotes a positive weighting function. This objective, hence only requires samples from the original distribution $\vxhat_0 \sim p(\vxhat)$.

\section{Methods}
\label{sec:methods}

\begin{figure}
    \centering
    \includegraphics[width=0.5\textwidth]{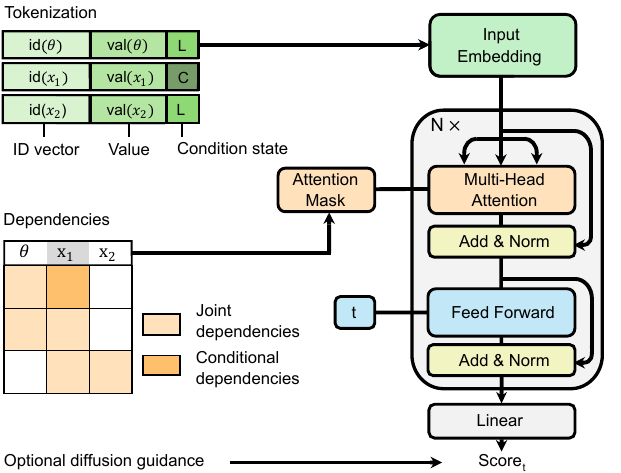}
    \caption{Simformer architecture. All variables (parameters and data) are reduced to a token representation which includes the variables' identity, the variables' value (val) as well as the conditional state (latent (L) or conditioned (C)). This sequence of tokens is processed by a transformer model; the interaction of variables can be explicitly controlled through an attention mask. The transformer architecture returns a score that is used to generate samples from the score-based diffusion model and can be modified (e.g. to guide the diffusion process).}
    \label{fig:fig1b}
    \vspace{-0.5cm}
\end{figure}

The Simformer is a probabilistic diffusion model that uses a transformer to estimate the score (\citet{weilbach2023graphically, hatamizadeh2023diffit, peebles2023scalable}, Fig.~\ref{fig:fig1b}). Unlike most previous approaches for simulation-based inference, which employ conditional density estimators to model either the likelihood or the posterior, the Simformer is trained on the \emph{joint} distribution of parameters and data $p(\vtheta, \vx) =: p(\vxhat)$. The Simformer encodes parameters and data (Sec.~\ref{subsec:tokenization}) such that arbitrary conditional distributions of the joint density (including posterior and likelihood) can still be sampled efficiently. The Simformer can encode known dependencies in the attention mask of the transformer (Sec.~\ref{subsec:dependencies}) and thereby ensures efficient training of arbitrary conditionals (Sec.~\ref{subsec:training}). Finally, the Simformer uses guided diffusion to produce samples given arbitrary constraints (Sec.~\ref{subsec:diffusion_guidance}).


\subsection{A Tokenizer for SBI}
\label{subsec:tokenization}

Transformers process sequences of uniformly sized vectors called tokens. Designing effective tokens is challenging and specific to the data at hand \cite{gu2022rethinking}. The tokenizer represents each variable as an identifier that uniquely identifies the variable, a representation of the value of the variable, and a condition state (Fig.~\ref{fig:fig1b}). The condition state is a binary variable and signifies whether the variable is conditioned on or not. It is resampled for every $(\vtheta, \vx) \in \mathbb{R}^d$ pair at every iteration of training. We denote the condition state of all variables as \( M_C \in \{0,1\}^d \). Setting $M_{C} = (0,\dots,0)$ corresponds to an unconditional diffusion model \cite{song2020score}, whereas adopting $M_{C}^{(i)} = 1$ for data and $M_{C}^{(i)} = 0$ for parameters corresponds to training a conditional diffusion model of the posterior distribution \cite{sharrock2022sequential,geffner2023compositional}. In our experiments, we uniformly at random sample either the masks for the joint, the posterior, the likelihood, or two randomly sampled masks (details in Appendix Sec.~\ref{sec:appendix_experimental_details}). To focus on specific conditional distributions, one can simply change the distribution of condition masks.

The Simformer uses learnable vector embeddings for identifiers and condition states, as proposed in 
 \citet{weilbach2023graphically}. In cases where parameters or data are functions of space or time, the node identifier will comprise a shared embedding vector and a random Fourier embedding of the elements in the index set.
Finally, specialized embedding networks, commonly used in SBI algorithms and trained end-to-end \cite{lueckmann2017flexible, chan2018exchangable, radev2020bayesflow}, can be efficiently integrated here by condensing complex data into a single token (e.g. we demonstrate this on a gravitational waves example in Appendix Sec.~\ref{sec:gw}). This reduces computational complexity but loses direct control over dependencies and condition states for individual data elements.

\subsection{Modelling dependency structures}
\label{subsec:dependencies}
For some simulators, domain scientists may have knowledge of (or assumptions about) the conditional dependency structures between parameters and data. For example, it may be known that all parameters are independent, or each parameter might only influence a single data value. The Simformer can exploit these dependencies by representing them in the attention mask $M_E$ of the transformer \citep{weilbach2023graphically}.
These constraints can be implemented as undirected (via a symmetric attention mask) or as directed dependencies (via a non-symmetric attention mask), that allow to enforce causal relations between parameters and observations. The latter, however, requires updating the mask if dependencies change i.e., due to conditioning \cite{webb2018faithful} (Fig.~\ref{fig:fig1b}, Appendix Sec.~\ref{sec:conditional_dependencies_appendix}).

A key advantage over masking weights directly \citep{germain2015made} is that the attention mask can be easily dynamically adapted at train or inference time, allowing to enforce dependency structures that are dependent on input values and condition state (details in Appendix Sec.~\ref{sec:properties_marg_cond}).
%
%
We note that the attention mask $M_E$ alone generally cannot ensure specific conditional independencies and marginalization properties in multi-layer transformer models. We describe the properties that can be reliably guaranteed and also explore how $M_E$ can be effectively employed to learn certain desired properties in Appendix Sec.~\ref{sec:properties_marg_cond}.

\subsection{Simformer training and sampling}
\label{subsec:training}


Having defined the tokenizer which processes every $(\vtheta, \vx)$ pair and the attention mask to specify dependencies within the simulator, the Simformer can be trained using denoising score-matching \citep{hyvarinen2005estimation, song2020score}:
We sample the noise level $t$ for the diffusion model uniformly at random and generate a (partially) noisy sample \( \hat{\mathbf{x}}_t^{M_C} = (1 - M_C) \cdot \hat{\mathbf{x}}_t + M_C \cdot \hat{\mathbf{x}}_0 \) i.e. variables that we want to condition on remain clean. The loss can then be defined as
\begin{align*}
\ell(\phi, M_C, t, \hat{\mathbf{x}}_0, \hat{\mathbf{x}}_t) =  \qquad   \qquad \qquad \qquad \qquad \qquad \quad & \\
(1-M_C)\cdot \left(s_\phi^{M_E}(\hat{\mathbf{x}}_t^{M_C}, t) - \nabla_{\hat{\mathbf{x}}_t}  \log p_t(\hat{\mathbf{x}}_t|\hat{\mathbf{x}}_0)\right),
\end{align*}
where $s_\phi^{M_E}$ denotes the score model equipped with a specific attention mask $M_E$. In expectation across noise levels $t$ and the data, this results in
\begin{equation*}
\mathcal{L}(\phi) = \mathbb{E}_{M_C, t, \hat{\mathbf{x}}_0, \hat{\mathbf{x}}_t}[ \left\lVert \ell(\phi, M_C, t, \hat{\mathbf{x}}_0, \hat{\mathbf{x}}_t) \right\rVert_2^2 ].
\end{equation*}
%
%
We note that to simplify notation, $M_E$ remains fixed here, but as stated in Sec.~\ref{subsec:dependencies}, it might depend on the condition state or input.

\begin{figure}[tp]
    \centering
    \includegraphics[width=0.5\textwidth]{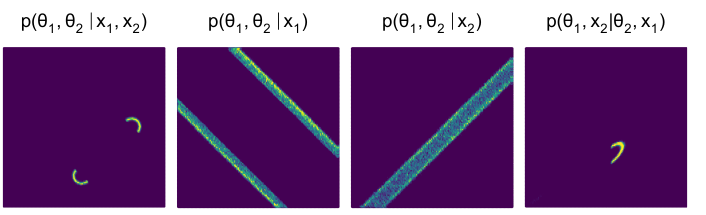}
    \caption{Examples of arbitrary conditional distributions of the Two Moons simulator, estimated by the Simformer.
    }
    \label{fig:fig2b}
    \vspace{-0.35cm}
\end{figure}

After having trained the Simformer, it can straightforwardly sample arbitrary conditionals (Fig.~\ref{fig:fig2b}). We draw samples from the noise distribution and run the reverse diffusion process on all unobserved variables, while keeping observed variables constant at their conditioning value \cite{weilbach2023graphically}.
Having access to all conditional distributions also allows us to combine scores and thereby perform inference for simulators with i.i.d.~datapoints \citep{geffner2023compositional}. Similarly, we can use other score transformations to adapt to other prior or likelihood configurations post-hoc (see Appendix Sec.~\ref{subsec:toy_appendix}).

\subsection{Conditioning on intervals with diffusion guidance}
\label{subsec:diffusion_guidance}

Guided diffusion makes it possible to sample from the generative model with an additional context $\mathbf{y}$, and has been used 
in tasks such as image inpainting, super-resolution, and image deblurring \cite{song2020score, chung2022improving}. It modifies the backward diffusion process to align it with a given context $\mathbf{y}$. Guided diffusion modifies the estimated score as
\begin{equation*}
s(\vxhat_t,t| \mathbf{y}) \approx s_\phi(\vxhat_t, t) + \nabla_{\vxhat_t} \log p_t(\mathbf{y}|\vxhat_t).
\end{equation*}
Various strategies for guiding the diffusion process have been developed, mainly differing in how they estimate $\nabla_{\vxhat_t} \log p_t(\mathbf{y}|\vxhat_t)$ \citep{ranzato2021diffusion,chung2023diffusion,jalal2021robust,song2022pseudoinverse,chung2022improving,bansal2023universal,lugmayr2022repaint}.

We here use diffusion guidance to be able to allow the Simformer to not only condition on fixed observations, but also on observation \emph{intervals} (or, similarly, intervals of the prior).
\citet{bansal2023universal} demonstrated that diffusion models can be guided by arbitrary functions. In that line, we use the following general formulation to guide the diffusion process:
$$s_\phi(\vxhat_t,t | c) \approx s_\phi(\vxhat_t,t) + \nabla_{\vxhat_t}  \log\sigma ( -s(t) c(\vxhat_t))$$
Here $\sigma$ denotes the sigmoid function, $s(t)$ is an appropriate scaling function satisfying $s(t)\rightarrow \infty$ as $t\rightarrow 0$, depending on the choice of SDE, and $c$ denotes a constraint function $c(\vxhat) \leq 0$. For example, to enforce an interval upper bound $u$, we use $c(\vxhat) = \vxhat - u$. We detail the algorithm used for guiding the diffusion process in Alg.~\ref{alg:general_guidance}.

\section{Results}
\label{sec:results}


\subsection{Benchmark tasks}
\label{sec:res_benchmark}

\begin{figure}
    \centering
    \includegraphics[width=0.5\textwidth]{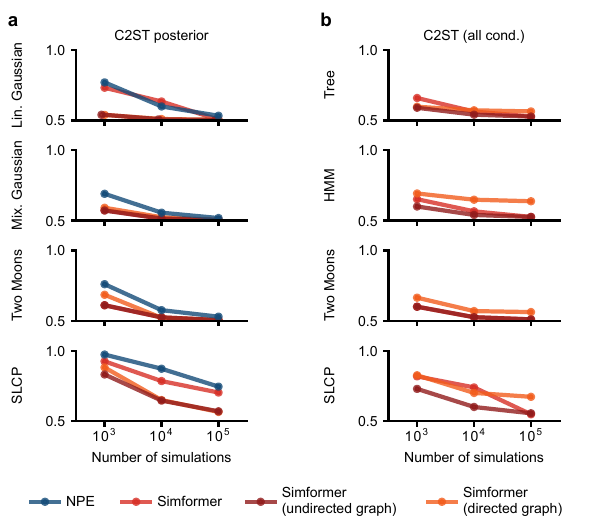}
    
    \caption{
    Simformer performance on benchmark tasks. The suffices "undirected graph" and "directed graph" denote Simformer variants with structured attention based on the respective graphical models.  \textbf{(a)} Classifier Two-Sample Test (C2ST) accuracy between Simformer- and ground-truth posteriors. \textbf{(b)} C2ST between arbitrary Simformer-conditional distributions and their ground truth.}
    \label{fig:fig2}
\end{figure}

We evaluated performance in approximating posterior distributions across four benchmark tasks \cite{lueckmann2021benchmarking}. For each task, samples for ten ground-truth posteriors are available (Appendix Sec.~\ref{sec:appendix_tasks}), and we assessed performance as classifier two-sample test (C2ST) accuracy to these samples. Here,  a score of 0.5 signifies perfect alignment with the ground truth posterior, and 1.0 indicates that a classifier can completely distinguish between the approximation and the ground truth. All results are obtained using the Variance Exploding SDE (VESDE); additional results using the Variance Preserving SDE (VPSDE) can be found in Appendix Sec.~\ref{sec:appendix_additional_results}. See Appendix Sec.~\ref{sec:appendix_experimental_details} for details on the parameterization.

Across all four benchmark tasks, the Simformer outperformed neural posterior estimation (NPE), even when the Simformer used a dense attention mask (Fig.~\ref{fig:fig2}a). The only exception was the Gaussian linear task with 10k simulations; we show an extended comparison with NRE and NLE in Appendix Fig.~\ref{fig:bm_vesde}, results with VPSDE in Appendix Fig.~\ref{fig:bm_vpsde}).  Incorporating domain knowledge into the attention mask of the transformer led to further improvements in the accuracy of the Simformer, particularly in tasks with sparser dependency structures, such as the Linear Gaussian (fully factorized) and SLCP (4 i.i.d.~observations). Averaged across all benchmark tasks and observations, the Simformer required about 10 times fewer simulations than NPE, leading to a vast reduction of computational cost for amortized inference.

Next, we evaluated the ability of the Simformer to evaluate arbitrary conditionals. Arbitrary parameter and data conditions often vastly differ from the form of the posterior distribution, leading to a challenging inference task (Fig.~\ref{fig:fig2b}). We performed inference on two of the benchmark tasks and established two new tasks with particularly interesting dependencies (Tree and HMM, details in Appendix Sec.~\ref{sec:appendix_tasks}). For each of the tasks, we generated ground truth posterior samples with Markov-Chain Monte-Carlo on 100 randomly selected conditional or full joint distributions. We found that, despite the complexity of these tasks, Simformer was able to accurately model all conditionals across all tasks (Fig.~\ref{fig:fig2}b). We note that training solely on the posterior mask does not enhance performance relative to learning all conditional distributions (Appendix Sec.~\ref{sec:appendix_additional_results}). Further, Simformer is well calibrated (Appendix Fig.~\ref{fig:cov_npe}, Fig.~\ref{fig:cov_sim}, Fig.~\ref{fig:cov_sim_und}, Fig.~\ref{fig:cov_sim_dir}) and, in most cases, also superior with respect to the loglikelihood (Appendix Fig.~\ref{fig:nll}).



\subsection{Lotka-Volterra: Inference with unstructured observations}

Many measurements in science are made in an unstructured way. For example, measurements of the populations of prey and predator species in ecology might not always be made at the same time points, and even the number of observations that were made might differ between species. To demonstrate that Simformer can deal with such `unstructured’ datasets, we applied the method to the ecological Lotka-Volterra model~\cite{lotka1925elements, volterra1926fluctuations}. The Lotka-Volterra model is a classic representation of predator-prey dynamics and is characterized by four global parameters, which govern the growth, hunting, and death rates of prey and predator. These populations evolve over time, guided by a set of coupled ordinary differential equations with Gaussian observation noise (details in Sec.~\ref{sec:appendix_tasks}).
\begin{figure}[h]
    \vspace{0.15cm}
    \centering
    \includegraphics[width=0.5\textwidth]{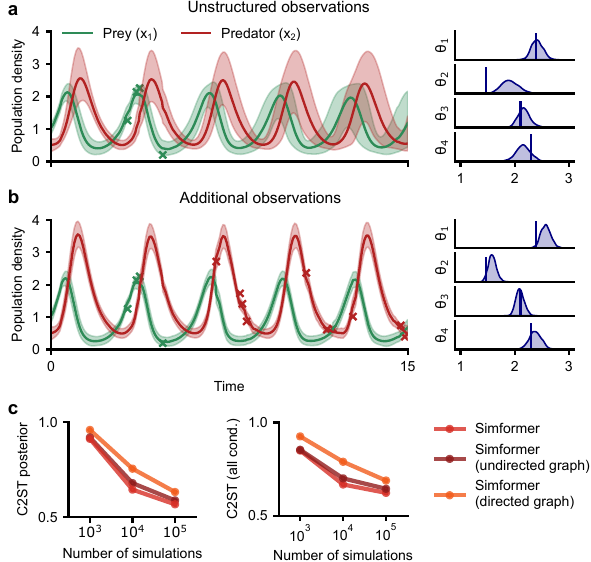}
    \caption{
    Inference with unstructured observations in the Lotka-Volterra model.
    \textbf{(a)} Posterior predictive (left) and posterior distribution (right) based on four unstructured observations of the prey population density (green crosses), using Simformer with $10^5$ simulations. True parameters in dark blue. \textbf{(b)} Same as (a) with nine additional observations of the predator population density. \textbf{(c)} C2ST-performance in estimating arbitrary conditionals (right) or the posterior distribution (left) using the C2ST metric.}
    \label{fig:fig3}
\end{figure}
We note that, unlike \citet{lueckmann2021benchmarking}, we perform inference for the \emph{full} time-series and do not rely on summary statistics.

We trained Simformer on $10^5$ simulations and, after training, generated several synthetic observations. The first of these observations contained four measurements of the prey population, 
placed irregularly in time (green crosses in Fig.~\ref{fig:fig3}a).

Using Simformer, we inferred the posterior distribution given this data. We found that the ground truth parameter set was indeed within regions of high posterior probability, and the Simformer posterior closely matched the ground truth posterior generated with MCMC (Fig.~\ref{fig:fig3}c, Appendix Sec.~\ref{sec:appendix_tasks}).
We then used the ability of Simformer to sample from arbitrary conditional distribution to simultaneously generate posterior and posterior predictive samples without additional runs of the simulator. The posterior predictives of Simformer capture data and uncertainty in a realistic manner (Fig.~\ref{fig:fig3}a).

As a second synthetic observation scenario, we used nine additional observations of the predator population, also irregularly placed in time (Fig.~\ref{fig:fig3}b). As expected, including these measurements reduces the uncertainty in both the posterior (Fig.~\ref{fig:fig3}b, right) and posterior predictive distributions (Fig.~\ref{fig:fig3}b left, posterior predictive again generated by the Simformer).


\subsection{SIRD-model: Inference in infinite dimensional parameters}
\label{subsec:sir}

Next, we show that Simformer can perform inference on functional data, i.e., $\infty$-dimensional parameter spaces, and that it can incorporate measurements of a subset of parameters into the inference process. In many simulators, parameters of the system may depend on time or space, and amortized inference methods should allow to perform parameter inference at \emph{any} (potentially infinitely many) points in time or space. We will demonstrate the ability of Simformer to solve such inference tasks in an example from epidemiology, the Susceptible-Infected-Recovered-Deceased (SIRD) model \cite{kermack1927contribution}. 

The SIRD simulator has three parameters: recovery rate, death rate, and contact rate. To simplify the inference task, these parameters are sometimes assumed to be constant in time, but treating the parameters as time-dependent allows to incorporate factors such as social distancing measures, public health interventions, and natural changes in human behavior \citep{chen2020sir,schmidt2021sir}. This is in contrast to \citet{lueckmann2021benchmarking}, which only considered a two-parameter SIR variant on a discrete-time grid. To demonstrate that Simformer can deal with a mixture of time-dependent and constant-in-time parameters, we assumed that the contact rate varied over time, whereas the recovery and death rate where constant in time.

\begin{figure}[t]
    \centering
    \includegraphics[width=0.5\textwidth]{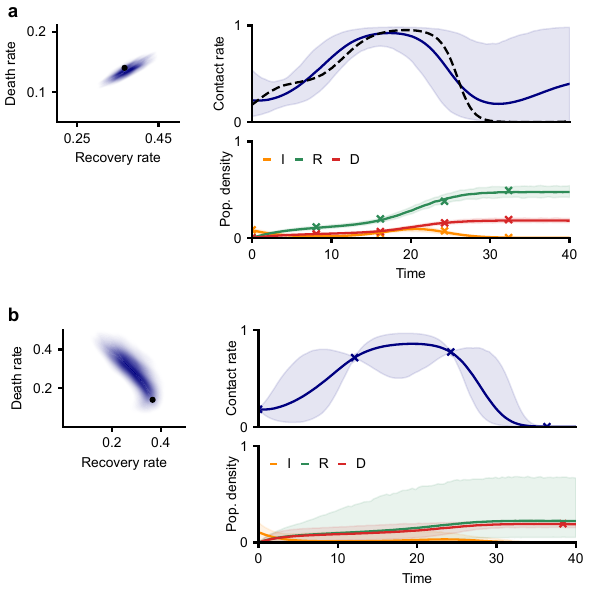}
    \caption{
    Inference of $\infty$-dim parameter space in the SIRD model.
    \textbf{(a)} Inferred posterior for global parameters (upper left) and time-dependent local parameters (upper right) based on five observations (crosses) of infected (I), recovered (R), and death (D) population densities. The black dot and dashed line indicate the true parameter, bold lines indicate the mean, and shaded areas represent \(99\%\) quantiles.
    \textbf{(b)} Inference with parameter measurements and a single measurement of fatalities.
    }
    \label{fig:fig4}
\end{figure}

\begin{figure*}
    \centering
    \includegraphics[width=\textwidth]{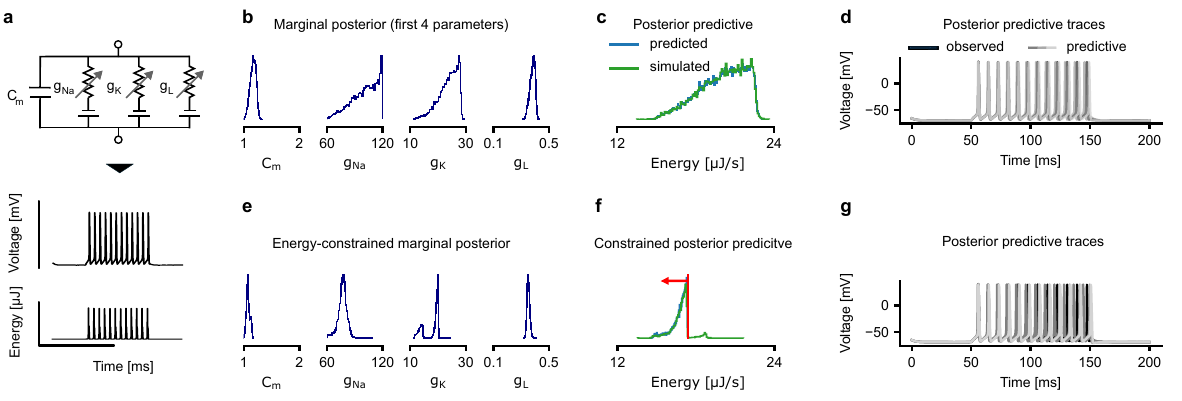}
    \caption{
    Inference in the Hodgkin-Huxley model.
    \textbf{(a)} Model schematic, observed voltage trace, and associated energy consumption. 
    \textbf{(b)} Marginals of inferred posterior for four parameters.
    \textbf{(c)} Posterior predictive energy consumption from Simformer (blue) and from simulator outputs (green). 
    \textbf{(d)} Posterior predictive samples from the posterior in (c) using the simulator.
    \textbf{(e)} Marginals of inferred energy constrained posterior for four parameters.
    \textbf{(f)} Posterior predictive energy consumption from Simformer (blue) and from simulator outputs (green). Energy constraint as red line.
    \textbf{(g)} Posterior predictive samples from posterior in (e) using the simulator.
}
    \label{fig:fig5}
\end{figure*}

We generated synthetic measurements from infected, recovered, and deceased individuals at irregularly spaced time points and applied the Simformer to estimate the posterior distribution of parameters. The Simformer estimated realistic death and recovery rates and successfully recovers a time-dependent contact rate that aligns with ground truth observations (Fig.~\ref{fig:fig4}a). Indeed, as measurements of infections tend towards zero (around timestamp 25, Fig.~\ref{fig:fig4}a, orange), the Simformer-posterior for the contact rate increases its uncertainty. This is expected, as we cannot obtain conclusive insights about the contact rate in scenarios with negligible infections. Additionally, as we already demonstrated on the Lotka-Volterra task, the ability of the Simformer to sample any conditional distribution allows us to generate posterior predictive samples without running the simulator. These samples closely match the observed data, further demonstrating the accuracy of the Simformer.

Next, we demonstrate that the Simformer can accurately sample parameter-conditioned posterior distributions (Fig.~\ref{fig:fig4}b). We generated a synthetic observation consisting of four measurements of the time-dependent contact rate parameter and a single measurement of infected people. The resulting Simformer-posterior closely aligns with the parameter measurements, and its posterior predictives are aligned with the data. We evaluate the performance quantitatively by computing the expected coverage, which verified that the conditional distributions estimated by Simformer are indeed well-calibrated (see Fig. \ref{fig:cov_sir}).

Overall, these results demonstrate that the Simformer can tackle function-valued parameter spaces and that its ability to sample arbitrary conditionals allows the incorporation of parameter measurements or assumptions into the inference procedure.

\subsection{Hodgkin-Huxley model: Inference with observation intervals}
\label{subsec:hh}

Finally, we demonstrate that the Simformer can perform inference in a highly nonlinear model and that it can constrain the parameters to observation \emph{intervals} with guided diffusion. For example, in neuroscience, it is desirable to obtain parameter configurations conditioned to experimental voltage measurements but also restricted by constraints such as lowering the metabolic cost (energy) below a particular threshold. Such additional constraints can be formalized as observation \emph{intervals}.

We demonstrate the ability of Simformer to perform such inferences in an example from neuroscience, the Hodgkin-Huxley simulator \citep{hodgkin1952}. This simulator describes the time course of voltage along the membrane of neurons (Fig.~\ref{fig:fig5}a). The simulator has $7$ parameters and generates a noisy time series, which we reduced to summary statistics as in previous work \citep{gonccalves2020training}. In addition, we also record the metabolic cost consumed by the circuit and add it as an additional statistic (Appendix Sec.~\ref{sec:appendix_tasks}).

We first inferred the posterior distribution given only the summary statistics of the voltage (not the energy) with the Simformer, and we found that, consistent with prior work \citep{gonccalves2020training}, the posterior distribution has wide marginals for some parameters and narrow marginals for others (Fig.~\ref{fig:fig5}b). We then used Simformer's ability to sample arbitrary conditionals and generate posterior predictives for energy consumption (Fig.~\ref{fig:fig5}c). The posterior predictive distribution of Simformer closely matched the posterior predictive distribution obtained by running the simulator (Fig.~\ref{fig:fig5}cd), and the energy cost of different posterior samples varied significantly \citep{deistler2022energy}. 

To identify energy-efficient parameter sets, we then defined an observation \emph{interval} for the energy consumption (energy must be within the lowest 10\% quantile of posterior predictives), and we used Simformer with guided diffusion to infer the posterior given voltage summary statistics and this constraint on energy consumption. The additional constraint on energy consumption significantly constrained the parameters posterior, in particular the maximal sodium and potassium conductances (Fig.~\ref{fig:fig5}e). We generated posterior predictive samples from this new posterior (via Simformer and by running the simulation) and found that their energy consumption indeed lies below the desired threshold (Fig.~\ref{fig:fig5}f). Furthermore, the corresponding predictive voltage trace is still in agreement with observations (Fig.~\ref{fig:fig5}g). Additional details and results on guidance are in Appendix Sec.~\ref{sec:appendix_guidance} (e.g. Fig.~\ref{fig:guidance_bm} for benchmarks on the accuracy of guidance).

Overall, Simformer can successfully recover the posterior distribution of highly nonlinear simulators. Simformer can condition on exact observations but also, using guided diffusion, on nearly arbitrary constraints (see Appendix Fig.~\ref{fig:guidance_modifications}, Fig~\ref{fig:guidance_two_moons}).

\section{Discussion}

We developed the Simformer, a new method for simulation-based amortized inference. The Simformer outperforms previous state-of-the-art methods (NPE) for posterior inference and simultaneously estimates all other conditionals. On tasks with notable independent structures, Simformer can be (on average across tasks and observations), one order of magnitude more simulation-efficient if equipped with a proper attention mask. 
The Simformer is significantly more flexible than previous out-of-the box inference frameworks and allows us to perform inference in $\infty$-dimensional parameter spaces, on unstructured and missing data. The Simformer makes it possible to sample arbitrary (or specified) conditional distributions of the joint distribution of parameters and data, including posterior and likelihood, thereby providing an `all-in-one' inference method. These conditional distributions can be used to perform inference with parameter conditionals, or to obtain posterior predictive samples without running the simulator. Using diffusion guidance, one can also condition on intervals, which, e.g., can be used to modify the prior without the need for retraining. Overall, the Simformer is an accurate and highly flexible inference method that opens up new possibilities for amortized inference methods in science and engineering.


\paragraph{Related Work}
The Simformer is designed to solve a range of problems in simulation-based inference, but its backbone, a probabilistic diffusion model on top of a transformer architecture, has also been used for generative models of images \citep{peebles2023scalable, hatamizadeh2023diffit}, and the task of generating arbitrary conditionals has been explored in various other generative models \cite{ivanov2019variational,li2020acflow, strauss2021arbitrary,strauss2022posterior}. 
In addition, integrating structural knowledge about the inference tasks has been previously explored for discrete diffusion models or continuous normalizing flows \cite{weilbach2020structured,harvey2022flexible,weilbach2023graphically} and has also been explored for neural processes and meta-learning \citep{nguyen2022transformer,nguyen2023transformer, müller2023transformers, maraval2023end}.

%
%

The benefits of diffusion models for simulation-based inference have also been explored: \citet{sharrock2022sequential} demonstrated that diffusion models can improve inference performance, and \citet{geffner2023compositional} showed that score decomposition can be used to perform inference for i.i.d.~data. The usage of diffusion models in the Simformer inherits these benefits. \citet{dax2023flow} demonstrated that flow-matching can largely reduce the number of trainable parameters needed for accurate inference results. \citet{schmitt2023fuse} proposed multi-head attention for integrating heterogeneous data from diverse sources. \citet{rozet2023scorebased} use a score-based model to learn the joint distribution of a dynamical system, approximately restricting their network to the Markovian structure, and then use guidance to condition it on specific observations.

The Simformer overcomes many limitations of current amortized inference methods, several of which have previously been tackled separately:
First, \citet{chen2020sir,ramesh2022gatsbi,moss2023simulationbased} also estimated posteriors over parameters that depended on space, but they relied on predefined discretizations to do so. 
Second, \citet{dyer2023approximate} inferred the posterior distribution for irregularly sampled time series via approximate Bayesian computation, and \citet{radev2020bayesflow} amortized inference across a flexible number of i.i.d.~trials (without considering irregularly sampled data).
Third, \citet{wang2023missing} proposed an approach to infer the posterior when data is missing, achieved through data augmentation and employment of recurrent neural networks.
Forth, whereas the Simformer inherently returns likelihood, posterior, and all other conditionals, \citet{radev2023jana} and \citet{gloeckler2022variational} learned separate networks for the likelihood and posterior and investigated features unlocked by having access to both distributions, and \citet{deistler2022energy} used MCMC to sample parameter conditionals of the learned posterior.
Finally, \citet{rozet2021arbitrary} proposed to estimate arbitrary marginal distributions for neural ratio estimation, whereas the Simformer can be used to estimate all conditional distributions.
All of the above works tackle the respective problem in isolation, whereas the architecture of the Simformer allows us to overcome all of these limitations at once.

\paragraph{Limitations}

Our method inherits the limitations of transformers and diffusion models: Generating samples is slower than for NPE, which is typically based on normalizing flows that permit fast sampling \citep{greenberg2019automatic}, whereas we have to solve the reverse SDE. On the other hand, sampling is much faster than methods that rely on MCMC \citep{papamakarios2019sequential,hermans2020likelihood}. In our experiments, accurate inference is achievable with as few as 50 evaluation steps, leading to sampling times of a few seconds for 10k samples. Further improvements may be possible by adapting the model \cite{song2022denoising}, the underlying SDE \citep{albergo2023stochastic} or SDE solver for sampling~\cite{gonzalez2023seeds}. 

Moreover, unlike normalizing flows, transformer evaluations scale quadratically with the number of input tokens, presenting significant memory and computational challenges during training. To mitigate this, various strategies have been proposed \citep{lin2022survey}. Naturally, using a sparse attention mask (e.g. due to many independencies) can reduce computational complexity~\citep{jaszczur2021sparse,weilbach2023graphically}.

In this work, we focus on estimating all conditionals, a task that, within our framework, is roughly as complex as learning the joint distribution. In problems with a few parameters but high dimensional data (i.e. images or long time series), estimating the joint might be harder than just the posterior. In such cases, Simformer can simply be queried to target specific conditionals of interest (e.g., posterior and missing data posteriors, see Appendix Sec. \ref{sec:gw} for an example on gravitational waves).


Lastly, normalizing flows enable rapid and precise assessments of the log-probability for posterior (or likelihood) approximations. This efficiency facilitates their integration into MCMC frameworks and aids the computation of point estimates, such as the Maximum A Posteriori (MAP) estimate. The score-based diffusion model employed by the Simformer also allows to evaluate log-probabilities (of any conditional of the joint), but this requires solving the corresponding probability flow ODE, which presents a computational burden \citep{song2020score}. Fortunately, for tasks such as MAP computation or integrating the Simformer likelihood into an MCMC scheme, there's no need to frequently assess log-probabilities. Instead, the score function can be utilized for gradient ascent to optimize the MAP or to perform Langevin-MCMC sampling, seamlessly incorporating the Simformer likelihood with such MCMC methods. 

\paragraph{Conclusion} 
We developed the Simformer, a new method for amortized simulation-based inference. On benchmark tasks, it performs at least as well or better as existing methods that only target the posterior, although the Simformer estimates all conditional distributions. The Simformer is highly flexible and can jointly tackle multiple amortized inference tasks more effectively than previous methods.


\section*{Software and Data}
We used JAX~\citep{jax2018github} as backbone and hydra~\citep{Yadan2019Hydra} to track all configurations. Code to reproduce results is available at \url{https://github.com/mackelab/simformer}.
We use SBI \cite{tejerocantero2020sbi} for reference implementations of baselines.

\section*{Impact Statement} Simulation-based inference (SBI) holds immense promise for advancing science across various disciplines. Our work enhances the accuracy and flexibility of SBI, thereby allowing scientists to apply SBI to previously unattainable simulators and inference problems. However, it is crucial to acknowledge the potential for the application of our method in less desirable contexts. Careful consideration of ethical implications is necessary to ensure the responsible use of our method.

\section*{Acknowledgements}

%

%
This work was supported by the German Research Foundation (DFG) through Germany’s Excellence Strategy – EXC-Number 2064/1 – Project number 390727645, the German Federal Ministry of Education and Research (Tübingen AI Center, FKZ: 01IS18039A), the `Certification and Foundations of Safe Machine Learning Systems in Healthcare' project funded by the Carl Zeiss Foundation, and the European Union (ERC, DeepCoMechTome, 101089288). MG and MD are members of the International Max Planck Research School for Intelligent Systems (IMPRS-IS).
CW and FW acknowledge the support of the Natural Sciences and Engineering Research Council of Canada (NSERC), the Canada CIFAR AI Chairs Program, Inverted AI, MITACS, the Department of Energy through Lawrence Berkeley National Laboratory, and Google. This research was enabled in part by technical support and computational resources provided by the Digital Research Alliance of Canada Compute Canada (alliancecan.ca), the Advanced Research Computing at the University of British Columbia (arc.ubc.ca), Amazon, and Oracle.


\bibliography{main}
\bibliographystyle{icml2024}

\newpage
\appendix
\setcounter{figure}{0}
\renewcommand{\thefigure}{A\arabic{figure}}
\setcounter{section}{0}
\renewcommand{\thesection}{A\arabic{section}}

\onecolumn
\section*{\LARGE Appendix}

\section{Conditional and marginalization properties}
\label{sec:properties_marg_cond}

In this section, we want to clarify what independence structures are exactly imposed by the Simformer equipped with a specific attention mask at the target distribution ($t=0$) and intermediate marginals ($t>0$) (Appendix Sec.~\ref{sec:conditional_dependencies_appendix}). We further state what marginalization properties you can expect a priori and how to adapt the training procedure to additionally enforce certain marginalization constraints (Appendix Sec.~\ref{sec:marginalization_appendix}). We then discuss how to extend to include post-hoc adaption of prior or likelihood (Appendix Sec.~\ref{sec:post_hoc}) and demonstrate the content on a toy example (Appendix Sec.~\ref{subsec:toy_appendix}).

\subsection{Conditional dependencies}
\label{sec:conditional_dependencies_appendix}

\begin{wrapfigure}{r}{5cm}
    \centering
    \includegraphics{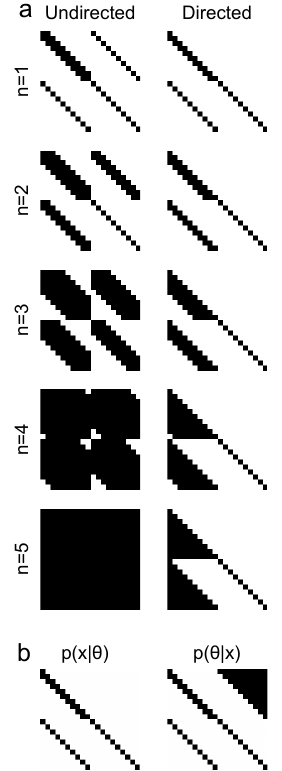}
    \caption{\textbf{(a)} Evolution of dependencies through $n=1,\dots,5$ transformer layers, given a constant attention mask for the HMM task ($n=1$). \textbf{(b)} Necessary adaption of the directed attention mask to faithfully capture conditional dependencies.}
    \label{fig:dependencies}
    \vspace{-30pt}
\end{wrapfigure}

We assume that the diffusion process (i.e. the underlying SDE) does not introduce any additional correlations, which is valid for VPSDE and VESDE. The attention mask, denoted by $M_E$, represents a graph $\mathcal{G}(\hat{\mathbf{x}}, M_E)$, with a total of $N$ vertices. We assume that $p(\vxhat)$ follows this graphical model. In this graph, if there exists a path from node $\hat{\mathbf{x}}_{i}$ to node $\hat{\mathbf{x}}_{j}$, then the transformer model $s_{\phi^*}^{M_E}$ is capable of attending $\hat{\mathbf{x}}_{j}$ to $\hat{\mathbf{x}}_{i}$, given it has enough layers. Conversely, the absence of such a path implies the transformer must estimate the score of $\hat{\mathbf{x}}_{i}$ independent of $\hat{\mathbf{x}}_{j}$. For an $l$-layer transformer, the matrix $D = \mathbb{I}(M_E^l > 0)$ succinctly represents all explicitly enforced conditional independencies, given a constant attention mask $M_E$. This is a classical result from graph theory i.e. that the n'th power of the adjacency matrix describes the number of walks from any node $i$ to any node $j$. The i'th row of this matrix delineates the variables upon which $\hat{\mathbf{x}}_{i}$ can attend and, therefore, potentially depend (see Fig.~\ref{fig:dependencies}a).

\paragraph{Dependencies at $\mathbf{t=0}$:} For an undirected, connected graph, all variables can depend on each other (given $l$ is large enough). This is a core argument by \citet{weilbach2023graphically} that an undirected graphical representation, given enough layers, is enough to faithfully represent all dependencies for any condition. Yet, this also diminishes any chance of correctly enforcing correct independencies beyond separating disconnected components. On the other hand, a directed acyclic graph will stay directed and acyclic. This property disallows modeling arbitrary dependencies, and this is why we have to dynamically adapt the mask to faithfully represent dependencies for arbitrary conditionals. We use the algorithm as proposed by \citet{webb2018faithful}, which returns a minimal amount of edges we have to add to the directed graph to faithfully represent present dependencies (under certain topological ordering constraints). This is shown in Figure \ref{fig:dependencies}b. As expected for modeling the likelihood, no additional edges have to be introduced. On the other hand, to model the posterior distribution, we have to insert additional edges into the upper right corner. Note that this mask is sufficient to represent dependencies with a 1-layer transformer and thus adds too many edges in general. For Gaussian linear tasks, where $M_E$ stands as an idempotent matrix (i.e. $M_E^2 = M_E$), resulting in $D = M_E$, this implies that all conditional independencies are correctly enforced, explaining the substantial enhancement in accuracy. Even if dependencies are not exactly enforced, as observed by both our work and \citet{weilbach2023graphically}, structured masks can enhance performance and computational complexity, particularly in the presence of notable independence structures.  It is important to note that these dependencies are what is enforced by the model, not what is necessarily learned.

\paragraph{Dependencies at $\mathbf{t > 0}$:} The score estimator does target the score of $p_t(\vxhat_t) = \int p(\vxhat_t|\vxhat) p(\vxhat)d\vxhat$. Notably, the imposed graphical model $\mathcal{G}$ is assumed to be valid at $p(\vxhat)$ but is generally invalid for $p_t(\vxhat_t)$. Directed graphical models are not closed under marginalization (beyond leave nodes)~\citep{maathuis2018handbook}. Undirected graphical models are closed but become fully connected in the case of diffusion models (for each connected component)~\citep{weilbach2020structured}. 
As highlighted by \citet{rozet2023scorebased}, one rationale for overlooking this concern is that for small values of $t$, indicating minimal noise, this assumption holds approximately true.  Further, as $t$ grows and noise accumulates, the mutual information between variables must decrease to zero by construction, implying that dependencies must be transformed from $M_E$ at $t=0$ to the identity mask $I$ at $t=T$. As also discussed before, the actual constraints imposed on the transformer score model is $D$, which does have an increased ``receptive field''. For undirected graphical models, this can be seen as equivalent to the notion of ``pseudo-markov blanckets'' introduced in \citet{rozet2023scorebased}. Given enough layers, this is sufficient to model all $p_t(\vxhat_t)$~\citep{weilbach2023graphically}, at the cost of explicitly enforcing known constraints at $t=0$. This is generally not true for the directed graphical model. It can faithfully represent all dependencies at time $t=0$, but can not necessarily exactly represent it at time $t > 0$.  Only if all connected components become autoregressive, it similarly can represent all dependencies. For further work, if it is desired to preserve the causal flow of information, it might be interesting to also consider more expressive graph representations. The class of ancestral graphs, for example, is closed under marginalization and can preserve the causal flow of information ~\citep{zhang2008causal}.

\subsection{Marginalization Properties}
\label{sec:marginalization_appendix}
Transformers, with their capability to process sequences of arbitrary lengths, present a compelling opportunity to exclude non-essential variables directly from the input. This is not merely a convenience but a method to reduce computational complexity, which is directly influenced by the length of the sequence. Therefore, omitting non-essential variables at the input stage is more efficient than removing them post hoc. Another unique ability, which is usually not possible for other models, is the possibility to compute marginal densities.

However, this selective exclusion comes with a specific prerequisite. The ability to drop variables is guaranteed only if, for any subset of variables $\{\hat{\mathbf{x}}_i, \hat{\mathbf{x}}_j, \ldots\}$, the dependency matrix $D$ satisfies $D_{ni} = 0, D_{nj} = 0, \ldots$ for all $n \neq i, j$. In simpler terms, this means that this subset of variables should not be able to attend to any outside variables. When examining the mask depicted in Fig. \ref{fig:dependencies}, it becomes evident that for a transformer with five layers and an undirected mask, we cannot safely omit any of the variables. Conversely, with a directed mask in place, we are able to safely sample $p(\vtheta)$ (first 10 elements) independently from $p(\vx)$ (last 10 elements).

Particularly in cases where the dependency matrix $D$ is densely populated, dropping out certain variables can change the output in an unexpected manner. This challenge can be addressed by training a transformer model to accurately estimate correct marginal distributions, which can be done using two techniques:
\begin{itemize}
    \item \textbf{Subsampling:} When we subsample \(\vxhat\) to a subset \(S\), resulting in \(\vxhat_{S}\), we effectively shift our target distribution to the specific marginal distribution \( p(\vxhat_{S}) \). This technique is particularly valuable for representing objects of infinite dimensionality. According to the Kolmogorov Extension Theorem, such objects can be characterized through their finite-dimensional marginal distributions. Therefore, our approach involves learning the distributions \( p(\hat{x}_{\tau_1}, \dots, \hat{x}_{\tau_N}) \) for a series of random samples \(\tau_1, \dots, \tau_N\) from the corresponding index set, typically represented by random time points. We can efficiently learn all finite-dimensional marginal distributions by randomly subsampling realizations of the process at these random time points. Additionally, it is particularly efficient because it reduces the sequence of variables during training. Importantly, this may necessitate modifying the attention mask, namely by ensuring that variables that were connected through a now-dropped node must be connected.
    \item \textbf{Modifying the attention mask:} Interestingly, altering the attention mask by a marginalization operation on the graph it represents is analogous to subsampling. For example, we may employ the identity mask to estimate all one-dimensional marginal distributions. The impact on the loss function can be reformulated as:
    $$ \mathcal{L}(\phi) = \mathbb{E}_{\vxhat_0, \vxhat_t}\left[ \| s_{\phi^*}^{I}(\vxhat_t) - s(\vxhat_0, \vxhat_t)\|_2^2 \right] 
    = \sum_{i=1}^d \mathbb{E}_{\vxhat_0, \vxhat_t}\left[ (s_{\phi^*}^{I}(\vxhat_t)^{(i)} - s(\vxhat_0, \vxhat_t)^{(i)})^2 \right].$$
    As each variable is processed independently, thus \(s_{\phi^*}^{I}(\vxhat_t)^{(i)} = s_{\phi^*}^{I}(\vxhat_t^{(i)})\) and for the family of SDEs (uncorrelated) we have \(s(\vxhat_0, \vxhat_t)^{(i)} = s(\vxhat_0^{(i)}, \vxhat_t^{(i)})\). Consequently,
    $$\mathcal{L}(\phi) = \sum_{i=1}^d \mathbb{E}_{\vxhat_0, \vxhat_t}\left[ (s_{\phi^*}^{I}(\vxhat_t^{(i)}) - s(\vxhat_0, \vxhat_t^{(i)}))^2 \right] = \sum_{i=1}^d \mathbb{E}_{\vxhat_0^{(i)}, \vxhat_t^{(i)}}\left[ (s_{\phi^*}^{I}(\vxhat_t^{(i)}) - s(\vxhat_0, \vxhat_t^{(i)}))^2 \right],$$
    This is essentially a sum of denoising score-matching losses for each one-dimensional marginal, verifying that it indeed aims to learn the correct marginal score. We can easily extend this result to other marginal distributions.
\end{itemize}

While we employed \textit{subsampling} in the Lotka Volterra and SIR example. We do provide an example of the latter technique in Sec. \ref{subsec:toy_appendix}.

\subsection{Post-hoc modifications}
\label{sec:post_hoc}
Altering the model configurations, such as employing different priors and likelihoods, is a consideration. \citet{elsemueller2023sensitivityaware} incorporated these modifications directly into their model. This is also possible here, but this method necessitates simulations across all configurations for training. Remarkably, our model allows a wide range of post-hoc adjustments after training on a single configuration, thus enabling it to represent a wide array of configurations. This flexibility is rooted in Bayes' rule, allowing for the decomposition of the score as
\begin{equation}
    \nabla_{\vtheta_t} \log p_t(\vtheta_t | \mathbf{x}_t) = \nabla_{\vtheta_t} \log p_t(\vtheta_t) + \nabla_{\vtheta_t} \log p_t(\mathbf{x}_t | \theta_t).
\end{equation}
Our model can estimate scores for the model it is trained on i.e. as described in Eq. \ref{eqn:toy_model}, but not for others. To address this limitation, we first can approximate
\begin{equation}
    \nabla_{\vtheta_t} \log p_t(\mathbf{x}_t | \vtheta_t) \approx s_\phi(\vtheta_t, t | \mathbf{x}_t) - s_\phi(\vtheta_t, t),
\end{equation}
and then adapt to a new family of model configurations using, for instance,
\begin{equation}
    \nabla_{\vtheta_t} \log p_t^{\alpha_1, \beta_1, \alpha_2, \beta_2}(\vtheta_t | \mathbf{x}_t) \approx \underbrace{\alpha_1 \cdot (s_\phi(\vtheta_t, t) + \beta_1)}_{\text{Prior change}} + \underbrace{\alpha_2 \cdot (s_\phi(\vtheta_t, t | \mathbf{x}_t) - s_\phi(\vtheta_t, t) + \beta_2)}_{\text{Likelihood change}}.
\end{equation}
This decomposition is also the main mechanism behind classifier-free guidance methods \cite{ho2022classifier}, which only act on the likelihood term. In general, $\alpha$ can temper the prior or likelihood, while $\beta$ can shift the location. Yet, the exact influence can only be inferred with the precise knowledge of the corresponding distribution at hand.

  \begin{wrapfigure}{r}{5cm}
     \centering   
     \includegraphics[width=5cm]{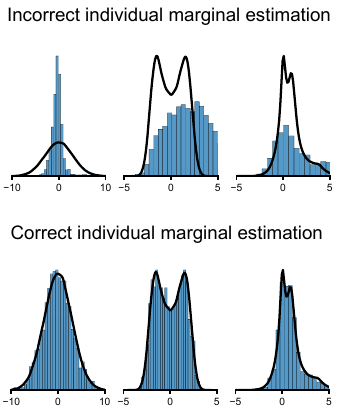}
     \caption{A model trained on a dense attention mask will predict the wrong marginal distribution without all other variables (top). A model trained also on the identity mask will provide correct marginals in the absence of all other variables (bottom)}
     \label{fig:correct_marginals}
     \vspace{-50px}
 \end{wrapfigure}

In a similar line, we are able to impose almost arbitrary constraints by manipulating the score accordingly. 
$$s_\phi(\vxhat_t,t | c) \approx s_\phi(\vxhat_t,t) + \nabla_{\vxhat_t} \sum_{i = 1}^K \log\sigma ( -s(t) c_i(\vxhat_t))$$
for a set of $K$ equations $c_i(\vxhat_t) \leq 0$, specifying a specific constraint, and a scaling function $s$. More details on the exact implementation and choices in Sec. \ref{sec:appendix_guidance}.

\subsection{Toy example}
\label{subsec:toy_appendix}

To demonstrate some of the above that we did not consider in the main paper, we consider a simple toy example of the form. 
 \begin{equation*}
 \label{eqn:toy_model}
     \theta \sim \mathcal{N}(0,3^2) \qquad x_1 \sim \mathcal{N}(2\cdot \sin(\theta), 0.5^2) \qquad x_2 \sim \mathcal{N}(0.1\cdot \theta^2 , 0.5 \cdot |x_1|)
 \end{equation*}

We train the Simformer using the following masks: (1) a dense mask for joint estimation, (2) an identity mask for accurate one-dimensional marginal estimation, and (3) two-dimensional marginal masks for precise two-dimensional marginal estimation. Indeed, in contrast to a model trained solely with a dense mask, our approach correctly estimates the marginals even in the absence of other variables, as shown in Fig.~\ref{fig:correct_marginals}. While both models can accurately capture the joint distribution (and consequently the marginals), this accuracy is contingent on receiving the complete sequence of variables as input.

\begin{figure}
    \centering
    \includegraphics[width=\textwidth]{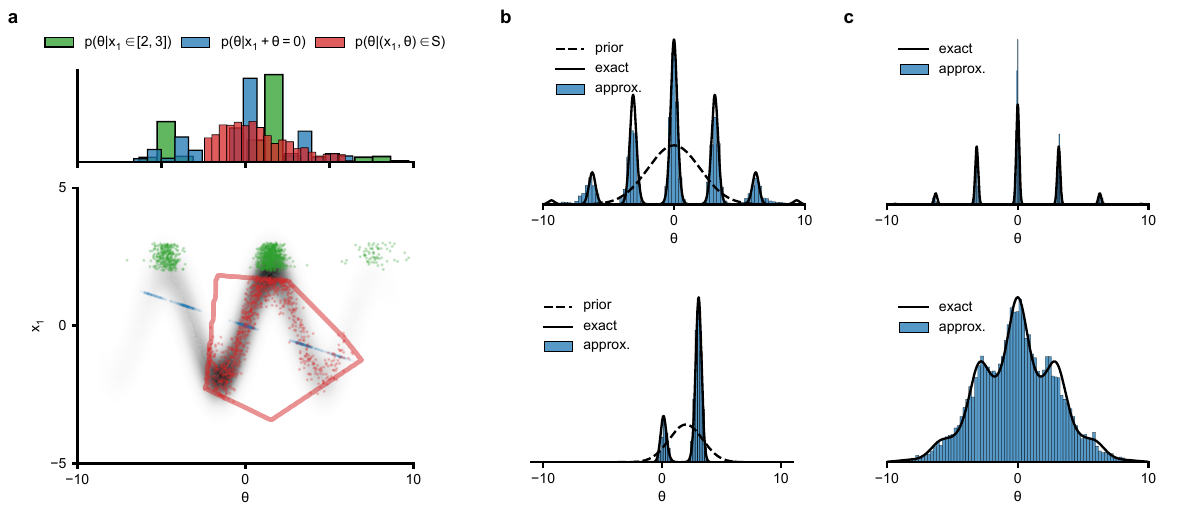}
    \caption{Illustration of the impact of post-hoc modifications on the 2d marginal posterior distribution for various model configurations, given the observation \( x_1 = 0 \). \textbf{(a)} Black shade shows ground-truth joint distributions. Scatter plots show samples with imposed constraints. \textbf{(b)} Posterior distribution with post-hoc modification to the prior i.e. increasing variance (top) or decreasing and shifting location. \textbf{(c)} Posteriors obtained by decreasing (top) or increasing (bottom) the variance of the likelihood}
    \label{fig:guidance_modifications}
\end{figure}

Next, we aim to impose certain constraints on a simplified version of diffusion guidance. Which are:
\begin{itemize}
    \item Interval: $c_1(x_1) =( x_1 -2)$ and $c_2(x_1) = (3-x_1)$.
    \item Linear: $c_1(x_1, \theta) = ( x_1 + \theta) $ and $c_2(x_1, \theta) =-( x_1 + \theta) $.
    \item Polytope: $c(x_1, \theta) = (A (x_1, \theta)^T - 1)$.
\end{itemize}
As visible in Fig. \ref{fig:guidance_modifications}, we indeed can enforce this constraint while predicting the correct associated $\theta$ distribution. 

Last but not least, we want to explore the capability to generalize to different generative models. In this example, with Gaussian distributions, affine transformations of approximate Gaussian scores will maintain their Gaussian nature, but we can alter the mean and variance. 

In the Gaussian scenario, we have
\begin{equation*}
    \nabla_x \log \mathcal{N}(x; \mu_0, \sigma_0^2) = \frac{x - \mu_0}{\sigma_0^2},
\end{equation*}
thus, to adjust this score to a specific mean \( \mu \) and variance \( \sigma^2 \), the appropriate choices would be
\begin{equation*}
    \alpha = \frac{\sigma_0^2}{\sigma^2}, \quad \text{and} \quad \beta = \frac{\mu - \mu_0}{\sigma_0^2}.
\end{equation*}

As demonstrated in Fig. \ref{fig:guidance_modifications}, these post hoc modifications indeed enable the computation of the posterior distribution for the same observation \( x_1 = 0 \) across diverse configurations. It is crucial to acknowledge, however, that these modifications have limitations, particularly if the changes are significantly divergent from the distributions of the initially trained model. This is evident in the figure, as increasing the prior variance works less well than decreasing it.

\newpage

\section{Experiment details}
\label{sec:appendix_experimental_details}

\subsection{Training and model configurations:}
\label{sec:appendix_model_details}

In our experiments, we adhere to the Stochastic Differential Equations (SDEs) as proposed by \citet{song2020score}, specifically the Variance Exploding SDE (VESDE) and the Variance Preserving SDE (VPSDE). These are defined as follows:

For VESDE:
\begin{equation}
    f_{\text{VESDE}}(x,t) = 0, \qquad g_{\text{VESDE}}(t) = \sigma_{min} \cdot \left( \frac{\sigma_{max}}{\sigma_{min}} \right)^t \cdot \sqrt{2\log \frac{\sigma_{max}}{\sigma_{min}}}
\end{equation}

For VPSDE:
\begin{equation}
    f_{\text{VPSDE}}(x,t) = -0.5 \cdot (\beta_{\text{min}} + t \cdot (\beta_{\text{max}} - \beta_{\text{min}})), \qquad g_{\text{VPSDE}}(t) = \sqrt{\beta_{\text{min}} + t \cdot (\beta_{\text{max}} - \beta_{\text{min}})}
\end{equation}

We set $\sigma_{\text{max}} = 15$, $\sigma_{\text{min}} = 0.0001$, $\beta_{\text{min}} = 0.01$, and $\beta_{\text{max}} = 10$ for all experiments. Both for the time interval $[1e-5,1.]$.

For implementing Neural Posterior Estimation (NPE), Neural Ratio Estimation (NRE), and Neural Likelihood Estimation (NLE), we utilize the sbi library \cite{tejerocantero2020sbi}, adopting default parameters but opting for a more expressive neural spline flow for NPE and NLE. Each method was trained using the provided training loop with a batch size of 1000 and an Adam optimizer. Training ceased upon convergence, as indicated by early stopping based on validation loss.

The employed transformer model features a token dimension of $50$ and represents diffusion time through a $128$-dimensional random Gaussian Fourier embedding. It comprises 6 layers and 4 heads with an attention size of $10$, and a widening factor of $3$, implying that the feed-forward block expands to a hidden dimension of $150$. For the Lotka-Volterra, SIR, and Hodgkin-Huxley tasks, we increased the number of layers to 8. Similar to the above, we used a training batch size of 1000 and an Adam optimizer.

\begin{figure}[tp]
    \centering
    \includegraphics[width=\textwidth]{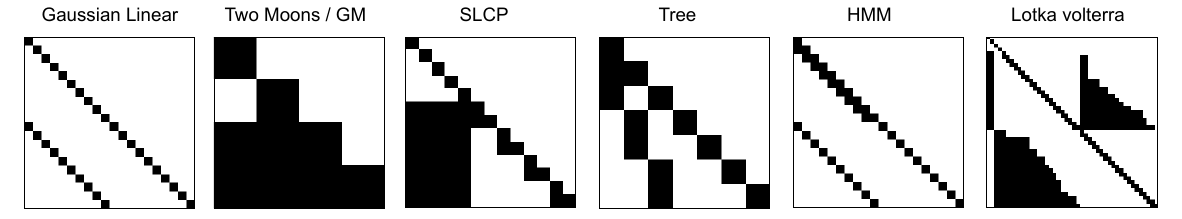}
    \caption{Directed \textit{base} masks for each of the tasks.The Lotka Volterra mask dynamically adapts to different input times, here just for randomly selected times.}
    \label{fig:masks}
\end{figure}

In all our experiments, we sampled the condition mask $M_C$ as follows: At every training batch, we selected uniformly at random a mask corresponding to the joint, the posterior, the likelihood or two random masks. The random masks were drawn from a Bernoulli distribution with $p=0.3$ and $p=7$. In our experiments, we found this to work slightly better than just random sampling and sufficiently diverse to still represent all the conditionals. The edge mask $M_E$ is chosen to match the generative process (see Fig.~\ref{fig:masks}). The undirected variant was obtained by symmetrization. Note that this is the only input we provide; additional necessary dependencies, e.g., due to conditioning, are algorithmically determined (see Sec. \ref{sec:conditional_dependencies_appendix}). 

For inference, we solved the reverse SDE using an Euler-Maruyama discretization. We use 500 steps by default; accuracy for different budgets is shown in Fig. \ref{fig:efficiency}.

\subsection{Tasks:}
\label{sec:appendix_tasks}

The tasks Gaussian Linear, Gaussian Mixture, Two Moons, and SLCP were used in \citet{lueckmann2021benchmarking}.

\paragraph{Gaussian Linear:} The prior for the parameter \(\vtheta\) is a normal distribution \( \mathcal{N}(0, 0.1\cdot \mathbf{I}) \). The data \(\vx\) given \(\vtheta\) is generated by a Gaussian distribution \( \mathcal{N}(\vx ; \vtheta,  0.1 \cdot \mathbf{I}) \). Both  \(\vtheta, \vx \in  \mathbb{R}^{10} \).

\paragraph{Gaussian Mixture}

This task, commonly referenced in Approximate Bayesian Computation (ABC) literature \cite{sisson2007sequential, BeaumontCornuet_09}, involves inferring the common mean of a mixture of two-dimensional Gaussian distributions with distinct covariances. The task is defined as follows. The prior for the parameters \(\vtheta\) is a uniform distribution, denoted as \( \mathcal{U}(-10, 10) \). The data \(\vx\) given \(\vtheta\) is modeled as a mixture of two Gaussian distributions:
    \[
        \vx | \vtheta \sim 0.5 \cdot \mathcal{N}(\vx ; \vtheta,  \mathbf{I}) + 0.5 \cdot \mathcal{N}(x; \vtheta, 0.01 \cdot \mathbf{I})
    \]
The parameter space \(\vtheta\) and the data space \(\vx\) are both in \( \mathbb{R}^2 \).

\paragraph{Two Moons}: The Two Moons task is designed to test inference algorithms in handling multimodal distributions 
~\cite{greenberg2019automatic}. The prior is a Uniform distribution \( U(\vtheta;-1, 1) \). The data \(\vx\) is generated from \(\vtheta\) as:
\begin{align*}
    \vx| \vtheta &= \begin{bmatrix} r \cos(\alpha) + 0.25 \\ r \sin(\alpha) \end{bmatrix} + \begin{bmatrix} -|\theta_1 + \theta_2|/\sqrt{2} \\ (-\theta_1 + \theta_2)/\sqrt{2} \end{bmatrix},
\end{align*}
where \( \alpha \sim \mathcal{U}(-\pi/2, \pi/2) \) and \( r \sim \mathcal{N}(0.1, 0.012) \). Leading to a dimensionality \(\vtheta \in  \mathbb{R}^2\), \(\vx \in  \mathbb{R}^2\).

To obtain reference samples for all possible conditionals, we run the following procedure:
\begin{itemize}
    \item We initialized $N$ Markov chains with samples from the joint distribution.
    \item We run 1000 steps of a random direction slice sampling algorithm.
    \item We run an additional 3000 steps of MHMCMC with step size of 0.01.
    \item Only the last samples of each chain were considered, yielding $N$ reference samples.
\end{itemize}
This procedure yielded samples in agreement with the reference posterior provided by \citet{lueckmann2021benchmarking} (C2ST $\sim 0.5$). Other conditionals did also look correct, but were not extensively investigated.

\paragraph{SLCP Task:} The SLCP (Simple Likelihood Complex Posterior) task is a challenging inference task designed to generate a complex posterior distribution \cite{papamakarios2019sequential, greenberg2019automatic, hermans2020likelihood, durkan2020contrastive}. The setup is as follows. The prior over $\vtheta$ is a uniform distribution \( \mathcal{U}(-3, 3) \). The data \(\vx\) given \(\vtheta\) is \(\vx = (\vx_1, \ldots, \vx_4)\), where each \(\vx_i \sim \mathcal{N}(\mathbf{\mu_{\theta}}, \mathbf{\Sigma_{\theta}}) \) with:
    \begin{align*}
        \mathbf{\mu_{\theta}} &= \begin{bmatrix} \theta_1 \\ \theta_2 \end{bmatrix}, \\
         \mathbf{\Sigma_{\theta}} &= \begin{bmatrix} \theta_3^2 & \tanh(\theta_5) \cdot \theta_3^2 \cdot \theta_4^2 \\ \tanh(\theta_5) \cdot \theta_3^2 \cdot \theta_4^2 & \theta_4^2 \end{bmatrix}.
    \end{align*}
Leading to a dimensionality of \(\vtheta \in  \mathbb{R}^5\), \(\vx \in  \mathbb{R}^8\).

To obtain reference samples for all possible conditionals, we run the following procedure:
\begin{itemize}
    \item We initialized $N$ Markov chains with samples from the joint distribution.
    \item We run 600 steps of a random direction slice sampling algorithm.
    \item We run an additional 2000 steps of MHMCMC with a step size of 0.1.
    \item Only the last samples of each chain was considered, yielding $N$ reference samples.
\end{itemize}
This procedure yielded samples in agreement with the reference posterior provided by \citet{lueckmann2021benchmarking} (C2ST $\sim 0.5$). Other conditionals did also look correct, but were not extensively investigated.

\paragraph{Tree: } This is a nonlinear tree-shaped task:
$$ \theta_0 \sim \mathcal{N}(\theta_0; 0, 1.) \qquad \theta_1 \sim \mathcal{N}(\theta_0; 1.) \qquad \theta_2 \sim \mathcal{N}(\theta_2; \theta_0, 1.).$$
Observable data is obtained through
$$ x_0 \sim \mathcal{N}(x_1; \sin(\theta_1)^2, 0.2^2) \quad x_1 \sim \mathcal{N}(0.1 \cdot \theta_1^2, 0.2^2) \quad x_2 \sim \mathcal{N}(x_2; 0.1\cdot \theta_2^2, 0.6^2) \quad x_3 \sim \mathcal{N}(x_3; \cos(\theta_2)^2; 0.1^2)$$
which leads to a tree-like factorization with highly multimodal conditionals.

To obtain reference samples for all possible conditionals, we run the following procedure:
\begin{itemize}
    \item We initialized $N$ Markov chains with samples from the joint distribution.
    \item We run 5000 steps of a HMC sampler.
    \item Only the last samples of each chain were considered, yielding $N$ reference samples.
\end{itemize}

\paragraph{HMM:} This is a task in which the parameters have a Markovian factorization.
$$ \theta_0 \sim \mathcal{N}(\theta_0; 0.,0.5^2) \qquad \theta_{i+1} \sim \mathcal{N}(\theta_{i+1}; \theta_i, 0.5^2)$$
for $i = 0,...,9$. Observations are generated according to $x_i = \mathcal{N}(x_i;\theta_i^2, 0.5^2)$, leading to a nonlinear hidden Markov model with bimodal correlated posterior and leading to a dimensionality of \(\vtheta \in  \mathbb{R}^{10}\), \(\vx \in  \mathbb{R}^{10}\).

To obtain reference samples for all possible conditionals, we run the following procedure:
\begin{itemize}
    \item We initialized $N$ Markov chains with samples from the joint distribution.
    \item We run 5000 steps of an HMC sampler.
    \item Only the last samples of each chain were considered, yielding $N$ reference samples.
\end{itemize}

\paragraph{Lotka Volterra}

The Lotka-Volterra equations, a foundational model in population dynamics, describe the interactions between predator and prey species \cite{volterra1926fluctuations, lotka1925elements}. This model is parameterized as follows: the prior is chosen to be a sigmoid-transformed Normal distribution, scaled to a range from one to three. Data then evolves according to the following differential equations:

\begin{equation}
    \begin{aligned}
        \frac{dx}{dt} &= \alpha x - \beta xy, \\
        \frac{dy}{dt} &= \delta xy - \gamma y.
    \end{aligned}
\end{equation}

Here, \( x \) and \( y \) represent the population sizes of the prey and predator species, respectively. The parameters \( \alpha, \beta, \gamma, \) and \( \delta \) are positive real numbers that describe the two species' interaction rates and survival rates. To each simulation, we add Gaussian observation noise with $\sigma=0.1$.

\paragraph{SIRD Model with Time-Dependent Contact Rate}

The SIRD (Susceptible, Infected, Recovered, Deceased) model extends the classical SIR framework by incorporating a Deceased (\(D\)) compartment. Similar models were explored by \citet{chen2020sir, schmidt2021sir}. This addition is crucial for modeling diseases with significant mortality rates. The dynamics of the SIRD model, considering a time-dependent contact rate, are governed by the following set of differential equations:

\begin{align*}
    \frac{dS}{dt} &= -\beta(t) SI, \\
    \frac{dI}{dt} &= \beta(t) SI - \gamma I - \mu I, \\
    \frac{dR}{dt} &= \gamma I, \\
    \frac{dD}{dt} &= \mu I.
\end{align*}

Here, \(S\), \(I\), \(R\), and \(D\) denote the number of susceptible, infected, recovered, and deceased individuals, respectively. The term \(\beta(t)\) represents the time-varying contact rate, while \(\gamma\) and \(\mu\) signify the recovery and mortality rates among the infected population, respectively.

Incorporating a time-dependent contact rate \(\beta(t)\) is pivotal for capturing the effects of public health interventions and societal behavioral changes over time. This feature is essential for accurately simulating the real-world dynamics of a disease's spread, particularly in the context of varying public health policies and community responses.

We impose a Uniform prior on the global variables, \(\gamma\) and \(\delta\), denoted as \(\gamma, \delta \sim \text{Unif}(0,0.5)\). For the time-dependent contact rate, we first sample  \(\hat{\beta} \sim \mathcal{G}(0,k)\) from a Gaussian process prior, with \(k\) representing an RBF kernel defined as \(k(t_1,t_2) = 2.5^2 \exp\left(-\frac{1}{2} \frac{\|t_1-t_2\|^2}{7^2}\right)\). This is further transformed via a sigmoid function to ensure \(\beta(t) \in [0,1]\) for all \(t\). Observational data is modeled with log-normal noise, characterized by a mean of \(S(t)\) and a standard deviation of \(\sigma = 0.05\).

\paragraph{Hodgkin-Huxley Model:}
In our study, we adhere to the implementation guidelines set forth by \citet{pospischil2008minimal} for the Hodgkin-Huxley model. The initial membrane voltage is established at $V_0 = -65.0\,\text{mV}$. Simulations are conducted over a duration of $200\,\text{ms}$, during which an input current of $4\,\text{mA}$ is applied in the interval between $50\,\text{ms}$ and $150\,\text{ms}$.

The rate functions are defined by the following equations:
\begin{align*}
    \alpha_m(V) &= 0.32 \times \frac{\text{efun}\left(-0.25(V - V_0 - 13.0)\right)}{0.25}, \\
    \beta_m(V) &= 0.28 \times \frac{\text{efun}\left(0.2(V - V_0 - 40.0)\right)}{0.2}, \\
    \alpha_h(V) &= 0.128 \times \exp\left(-\frac{(V - V_0 - 17.0)}{18.0}\right), \\
    \beta_h(V) &= \frac{4.0}{1.0 + \exp\left(-\frac{(V - V_0 - 40.0)}{5.0}\right)}, \\
    \alpha_n(V) &= 0.032 \times \frac{\text{efun}\left(-0.2(V - V_0 - 15.0)\right)}{0.2}, \\
    \beta_n(V) &= 0.5 \times \exp\left(-\frac{(V - V_0 - 10.0)}{40.0}\right)
\end{align*}
where \(\text{efun}(x) = \begin{cases} 1-\frac{x}{2} & \text{if } x < 1e-4 \\ \frac{x}{\exp(x) - 1.0} & \text{otherwise} \end{cases}\).

This formulation leads to the comprehensive Hodgkin-Huxley differential equations:
\begin{align*}
    \frac{dV}{dt} &= \frac{I_{\text{inj}}(t) - g_{\text{Na}} m^3 h (V - E_{\text{Na}}) - g_{\text{K}} n^4 (V - E_{\text{K}}) - g_{\text{L}} (V - E_{\text{L}})}{C_m} + 0.05\,dw_t, \\
    \frac{dm}{dt} &= \alpha_m(V) (1 - m) - \beta_m(V) m, \\
    \frac{dh}{dt} &= \alpha_h(V) (1 - h) - \beta_h(V) h, \\
    \frac{dn}{dt} &= \alpha_n(V) (1 - n) - \beta_n(V) n, \\
    \frac{dH}{dt} &= g_{\text{Na}} m^3 h (V - E_{\text{Na}}).
\end{align*}

Notably, there exist multiple methodologies for estimating energy consumption in neuronal models, as discussed by \citet{deistler2022energy}. In our approach, we opt to calculate energy consumption based on sodium charge, which can be converted into $\mu J/s$ as detailed by \citet{deistler2022energy}. For observational data, we employ summary features consistent with those used by \citet{gonccalves2020training}.

\newpage

\section{Additional experiments}
\label{sec:appendix_additional_results}

In Sec. \ref{sec:appendix_benchmark_extended}, we include additional experiments, i.e., investigating different SDEs, comparing to more methods, adding additional metrics, and reviewing efficiency. In Sec.~\ref{sec:gw}, we demonstrate target inference with embedding nets on a complex task for gravitational wave data. Finally, in Sec. \ref{sec:appendix_guidance}, we review how good guidance methods can compute arbitrary conditionals, as well as general constraints.

\subsection{Extended benchmark}
\label{sec:appendix_benchmark_extended}

\begin{figure}[H]
    \centering
    \includegraphics[width=\textwidth]{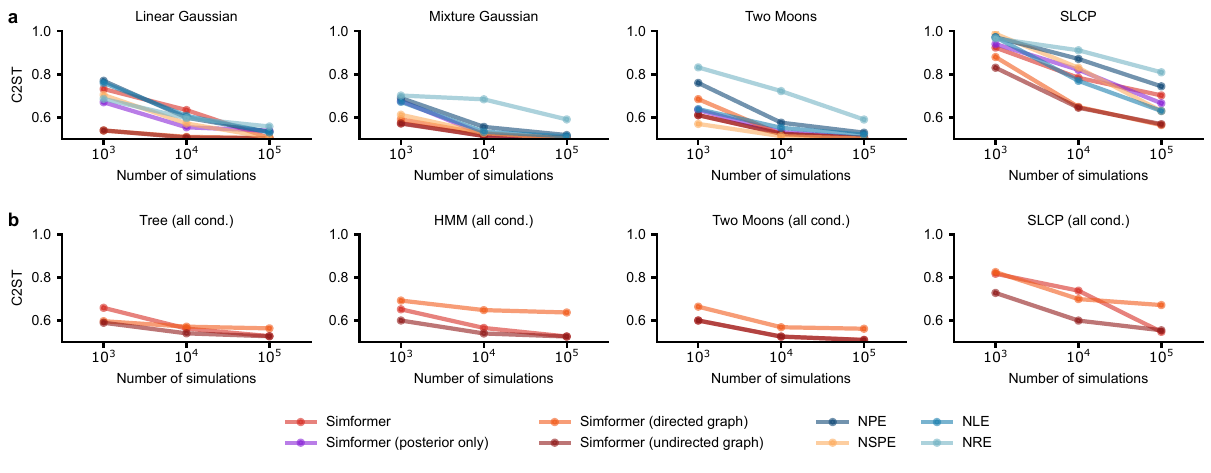}
    \caption{Extended benchmark results for the VESDE. In addition to NPE, we also run NRE, NLE, and NSPE. \textbf{(a)} Shows performance in terms of C2ST for SBIBM tasks. \textbf{(b)} Shows performance in terms of C2ST for all conditional distributions. }
    \label{fig:bm_vesde}
    \vspace{-1cm}
\end{figure}

\begin{figure}[H]
    \centering
    \includegraphics[width=\textwidth]{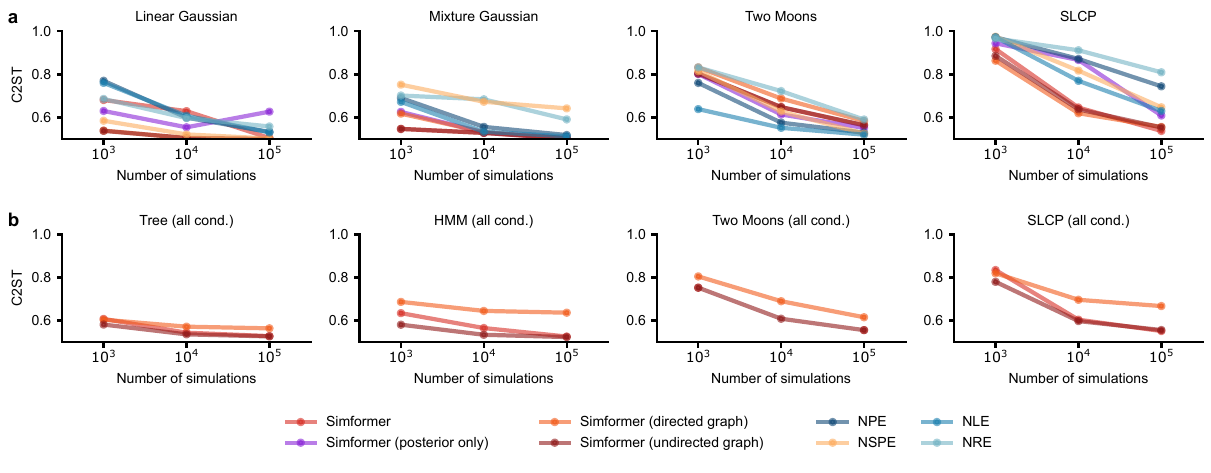}
    \caption{Extended benchmark results for the VPSDE. In addition to NPE, we also run NRE, NLE, and NSPE. \textbf{(a)} Shows performance in terms of C2ST for SBIBM tasks. \textbf{(b)} Shows performance in terms of C2ST for all conditional distributions.}
    \label{fig:bm_vpsde}
\end{figure}

\paragraph{Overview of benchmark results:}
Comprehensive benchmark results have been obtained for both the Variance Exploding SDE (VESDE) and the Variance Preserving SDE (VPSDE) models, as well as for several SBI methods. These methods include Neural Posterior Estimation (NPE) \cite{papamakarios2016fast}, Neural Likelihood Estimation (NLE) \cite{papamakarios2019sequential}, and Neural Ratio Estimation (NRE) \cite{hermans2020likelihood}. The outcomes of these benchmarks are depicted in Figure \ref{fig:bm_vesde} and Figure \ref{fig:bm_vpsde}. 

Furthermore, we have implemented a baseline Neural Posterior Score Estimation (NPSE) method \cite{sharrock2022sequential, geffner2023compositional}, where the score network is a conditional MLP in contrast to the transformer architecture. Additionally, a variant named the 'Simformer (posterior only)' was tested, in which the training focuses exclusively on the associated posterior masks, rendering its neural network usage akin to NPSE (up to different architectures). As expected, these two approaches do perform similarly. Furthermore, this shows that targeting all conditionals does not hurt (but can even improve) the performance even when evaluating the posterior only. 

\paragraph{Comparative performance of SDE variants:}
Overall, the different SDE variants exhibit comparably high performance, with some notable exceptions. Specifically, the VESDE model demonstrates superior performance in the Two Moons task, whereas the VPSDE model shows a slight edge in the SLCP task. 

\paragraph{Impact of training only on posterior masks:}
Interestingly, training solely on the posterior mask does not enhance performance relative to learning all conditional distributions. This observation confirms our initial hypothesis that the desired property of efficient learning of all conditionals is inherently 'free' in our framework. In cases like the SLCP, where the likelihood is relatively simple, there appears to be an added advantage in learning both the posterior and the likelihood distributions. Traditionally, likelihood-based methods such as NLE outperform direct posterior estimation techniques on this task. As the Simformer approach estimates both quantities jointly, it may benefit from this additional information.

\begin{figure}[H]
    \centering
    \includegraphics[width=\textwidth]{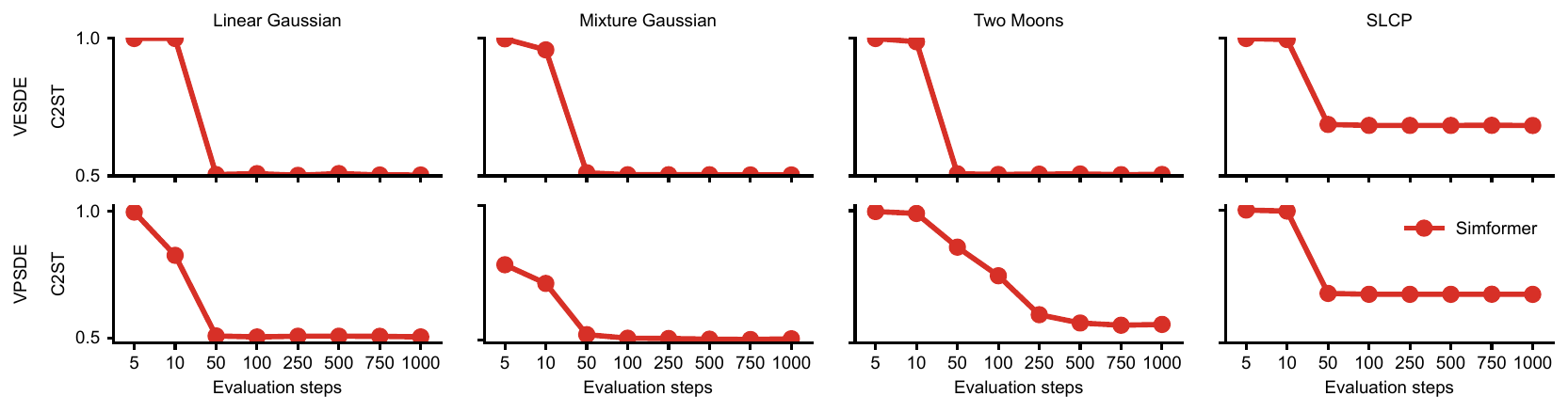}
    \caption{For all tasks as well as the VPSDE and VESDE, we show how the performance as measured in C2ST increases as we increase the evaluation steps to solve the reverse SDE. For all tasks, except Two Moons on the VPSDE, 50 evaluations are sufficient to reach best performance.}
    \label{fig:efficiency}
\end{figure}

\paragraph{Model evaluations for reverse diffusion:} In Figure \ref{fig:efficiency}, we illustrate how the C2ST varies with the number of model evaluations used in solving the reverse SDE. This variation is observed by examining different uniform discretizations of the time interval $[0, 1]$ with varying numbers of elements. Notably, the performance improvement of the method with an increasing number of evaluations is not gradual. Rather, there is a sharp transition from suboptimal to near-perfect performance when the number of evaluations exceeds 50. This finding is particularly favorable for diffusion models, as opposed to methods like NLE or Neural Ratio Estimation NRE, which necessitate a subsequent Markov Chain Monte Carlo (MCMC) run. It is important to note that these MCMC runs typically require significantly more than 50 evaluations, highlighting the efficiency of diffusion models in this context. This is especially important as transformer models are usually more expensive to evaluate than the network architectures used in NLE and NRE.

\begin{figure}[H]
    \centering
    \includegraphics[width=\textwidth]{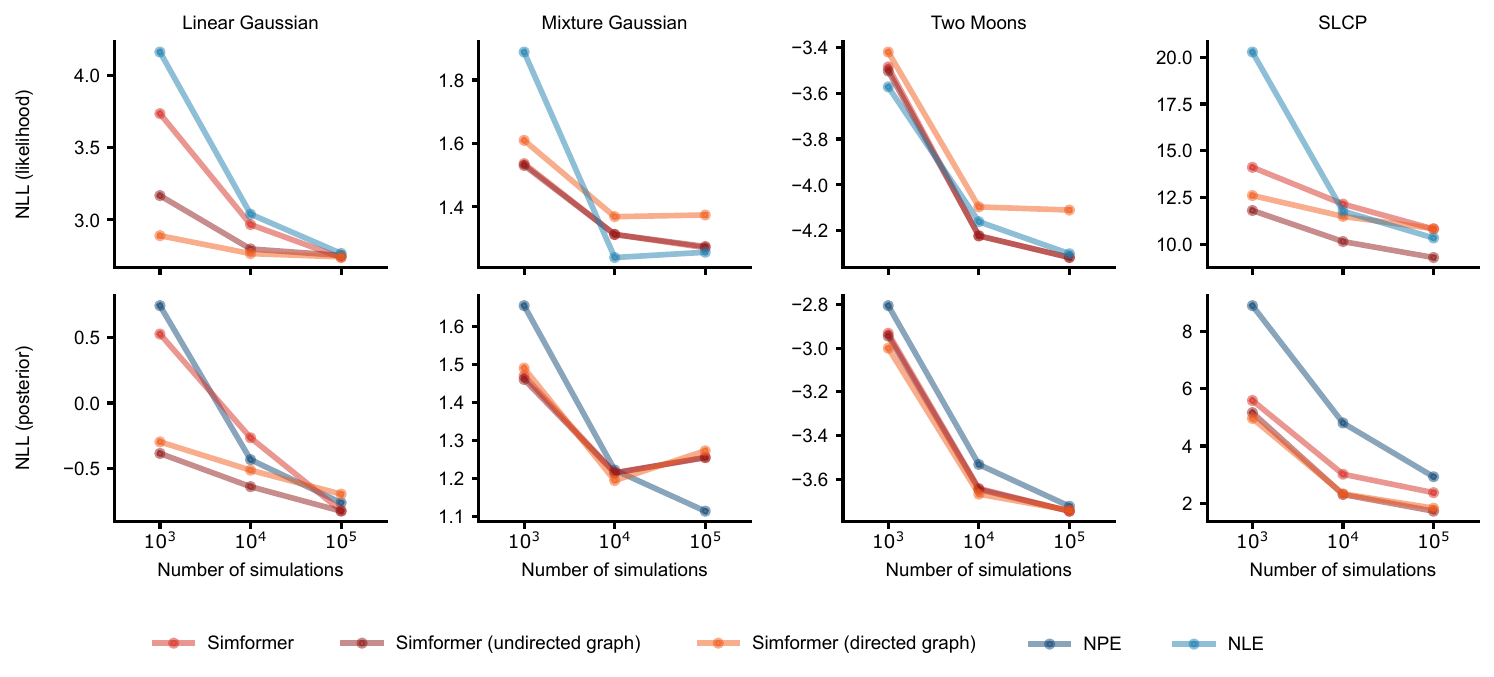}
    \caption{Average negative loglikelihood of the true parameter for NPE, NLE, and all Simformer variants. Evaluating both the likelihood (top row) and posterior (bottom row).}
    \label{fig:nll}
\end{figure}

\paragraph{Average negative loglikelihood:} The average negative loglikelihood (NLL) for the true posterior is a metric suitable for evaluation on an increased number of different observations~\cite{lueckmann2021benchmarking, hermans2022trust}. We evaluate the average on 5000 samples from the joint distribution. We did this for both the posterior and likelihood, as estimated by Simformer, and compared it to the corresponding NPE and NLE baseline. Note that NPE and NLE are trained to minimize the NLL, giving it a natural advantage. In contrast, Simformer only indirectly minimizes negative loglikelihood through the score-matching objective. Notably, to evaluate the loglikelihood for the Simformer, we have to use the probability flow ODE \cite{song2020score}. Hence, the loglikelihood is also based on the probability flow ODE, not the corresponding SDE formulation (which does not necessarily exactly agree for a finite number of steps). We show the corresponding result in Fig~\ref{fig:nll}. In most cases, the results agree with the C2ST evaluation (which only evaluates SDE sampling quality). However, in some cases NLE or NPE does perform better with respect to this metric. The difference is due to the discrepancy between SDE sampling and ODE log probability evaluation and the fact that Simformer is not trained to minimize loglikelihood, which is not necessarily at odds with producing good samples.

\begin{figure}[H]
    \centering
    \includegraphics[width=\textwidth]{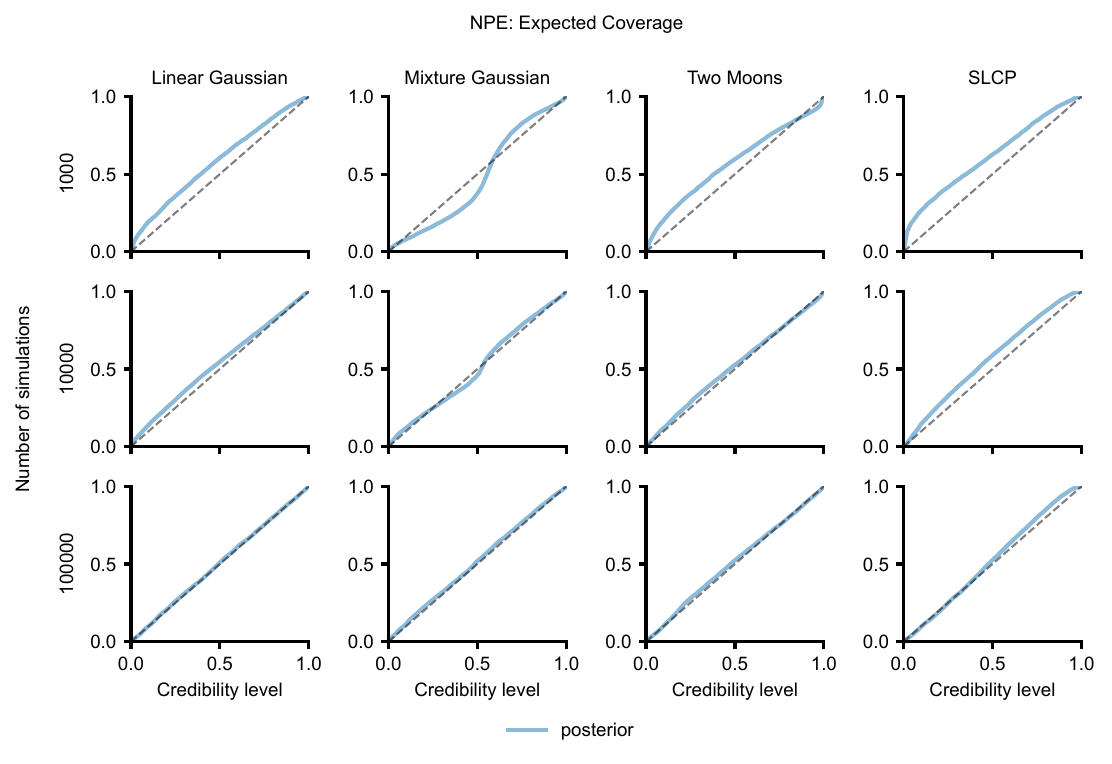}
    \caption{Calibration analysis for NPE using \textit{expected coverage}~\cite{hermans2022trust}. Each row corresponds to training simulation sizes of 1k, 10k, 100k. Each column represents a task.}
    \label{fig:cov_npe}
\end{figure}

\begin{figure}[H]
    \centering
    \includegraphics[width=\textwidth]{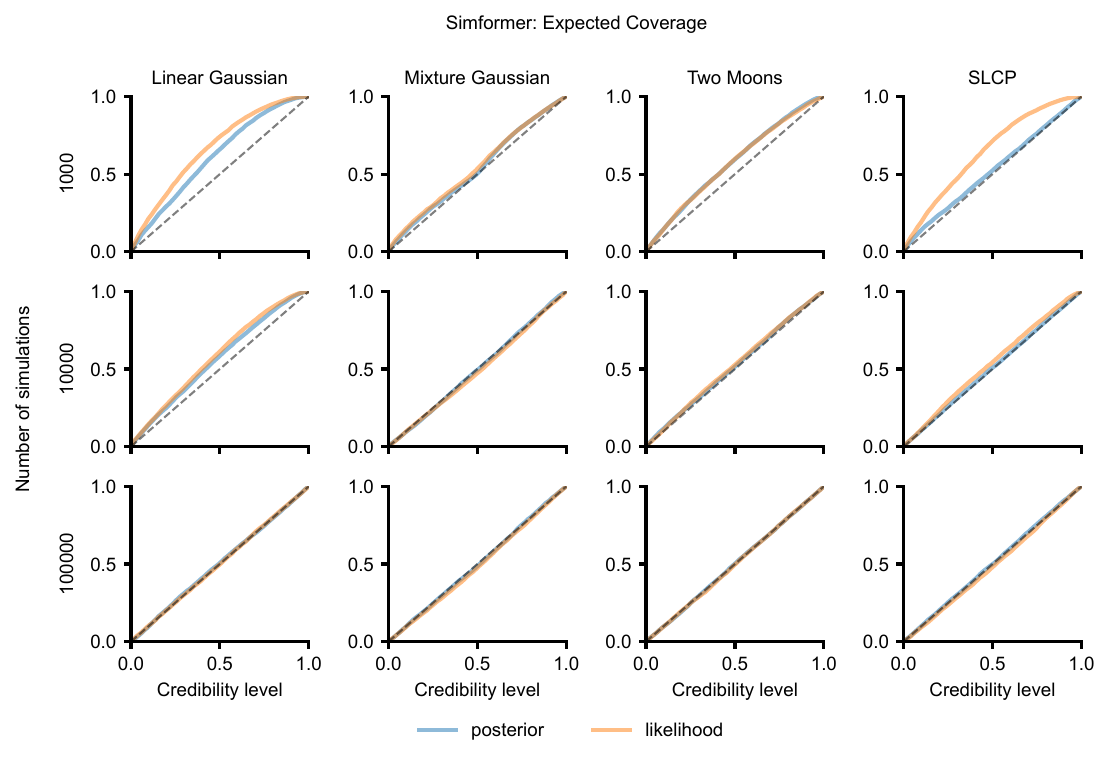}
    \caption{Calibration analysis for Simformer using \textit{expected coverage}~\cite{hermans2022trust}, both for the posterior and likelihood. Each row corresponds to training simulation sizes of 1k, 10k, 100k. Each column represents a task.}
    \label{fig:cov_sim}
\end{figure}

\begin{figure}[H]
    \centering
    \includegraphics[width=\textwidth]{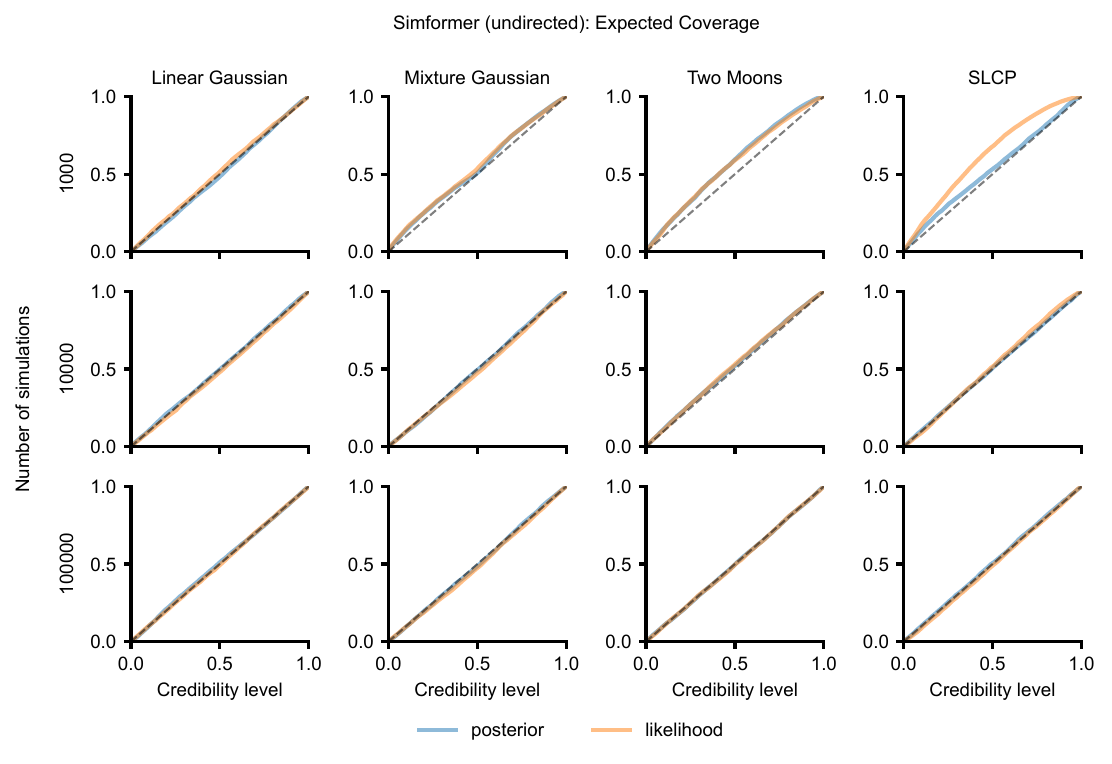}
    \caption{Calibration analysis for Simformer (undirected) using \textit{expected coverage}~\cite{hermans2022trust}, both for the posterior and likelihood. Each row corresponds to training simulation sizes of 1k, 10k, 100k. Each column represents a task.}
    \label{fig:cov_sim_und}
\end{figure}

\begin{figure}[H]
    \centering
    \includegraphics[width=\textwidth]{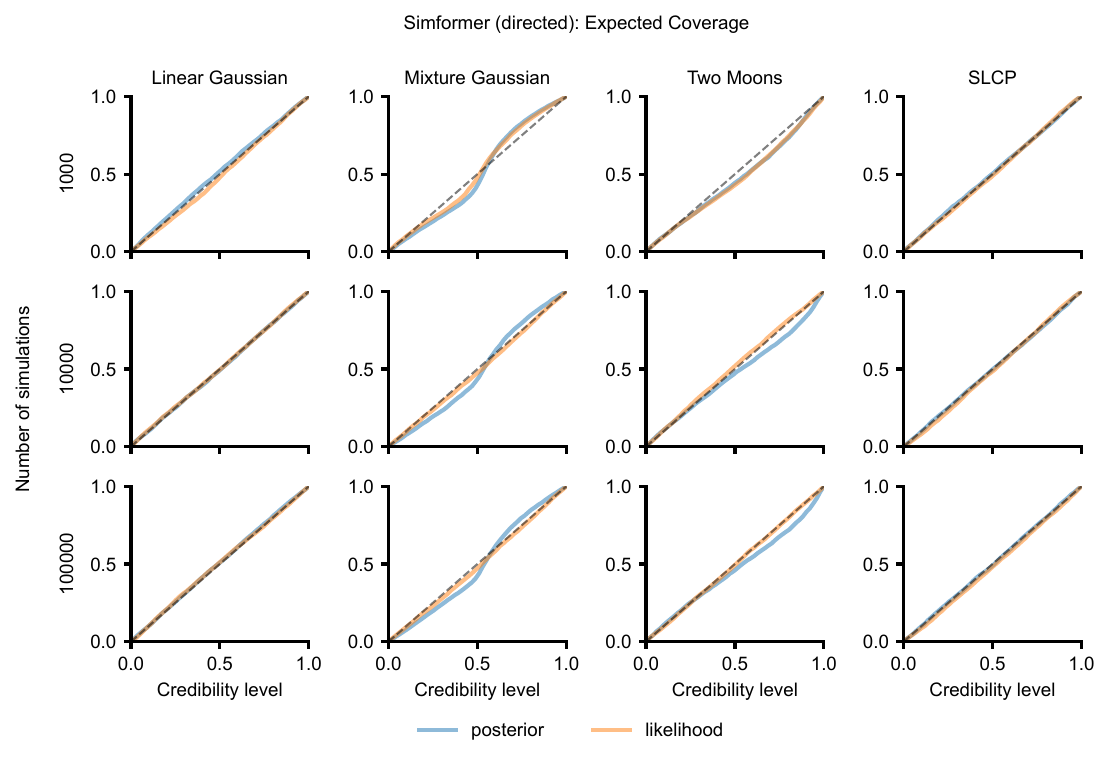}
    \caption{Calibration analysis for Simformer (directed) using \textit{expected coverage}, both for the posterior and likelihood. Each row corresponds to training simulation sizes of 1k, 10k, 100k. Each column represents a task.}
    \label{fig:cov_sim_dir}
\end{figure}

\paragraph{Calibration:} To check whether the distributions estimated by Simformer are well-calibrated, we performed an expected coverage analysis~\cite{hermans2022trust}, again both for the posterior and likelihood. Intuitively, this test checks whether the ground-truth parameter lies within the top $\alpha\%$ highest density region in $\alpha\%$ of all cases (which is what the true posterior must satisfy). The same analysis was performed for NPE as a reference (see Fig. \ref{fig:cov_npe}). In cases in which the likelihood is significantly easier to learn than the posterior (i.e., SLCP), we can observe that, indeed, the estimate of the simple likelihood becomes well calibrated earlier than the posterior (see Fig.~\ref{fig:cov_sim}, Fig.~\ref{fig:cov_sim_und}, Fig.~\ref{fig:cov_sim_dir}, upper right corner). Overall, Simformer is well-calibrated and, similar to NPE, tends to more \textit{conservative} approximations (coverage plots tend to be above the diagonal).

\begin{figure}[H]
    \centering
    \includegraphics[width=0.6\textwidth]{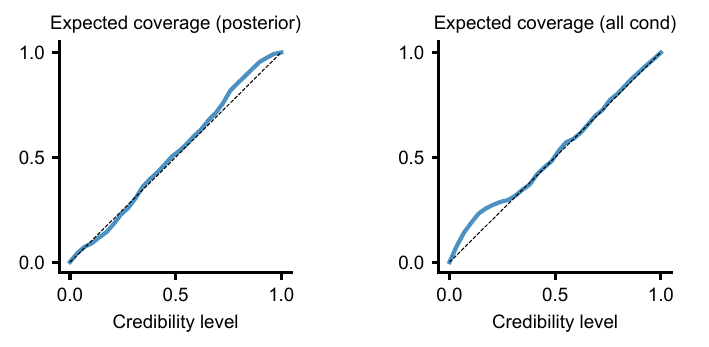}
    \caption{Calibration analysis for the SIR task using \textit{expected coverage}~\cite{hermans2022trust}. On the left, we evaluate the posterior (for randomly selected time points). On the right, we have the coverage for different conditional distributions (also for randomly selected time points).}
    \label{fig:cov_sir}
\end{figure}

We also perform a coverage analysis on the SIR task (Fig.~\ref{fig:cov_sir}). Note that because this model is nonparametric, there are infinitely many distributions we could evaluate (i.e. by selecting different times for observations or parameters). We opt to run an evaluation on 20 random time points for each time-dependent parameter (contact rate) or observation (S, I, D).

\newpage

\subsection{Targeteted inference and embedding nets}
\label{sec:gw}

In the main manuscript, we focus on estimating all conditionals of a certain task. However, in certain scenarios, it might simply not be wanted or way harder to do so. In this case, we can query Simformer to simply target only a subset of conditionals by restricting the number of condition masks $M_C$ to whatever conditionals we deem worth estimating. Secondly, in tasks were data is high dimensional, it becomes computationally demanding to consider each scalar as a variable. In this case, we should encode whole vectors into a single token.

\begin{figure}[H]
    \centering
    \includegraphics[width=0.88\textwidth]{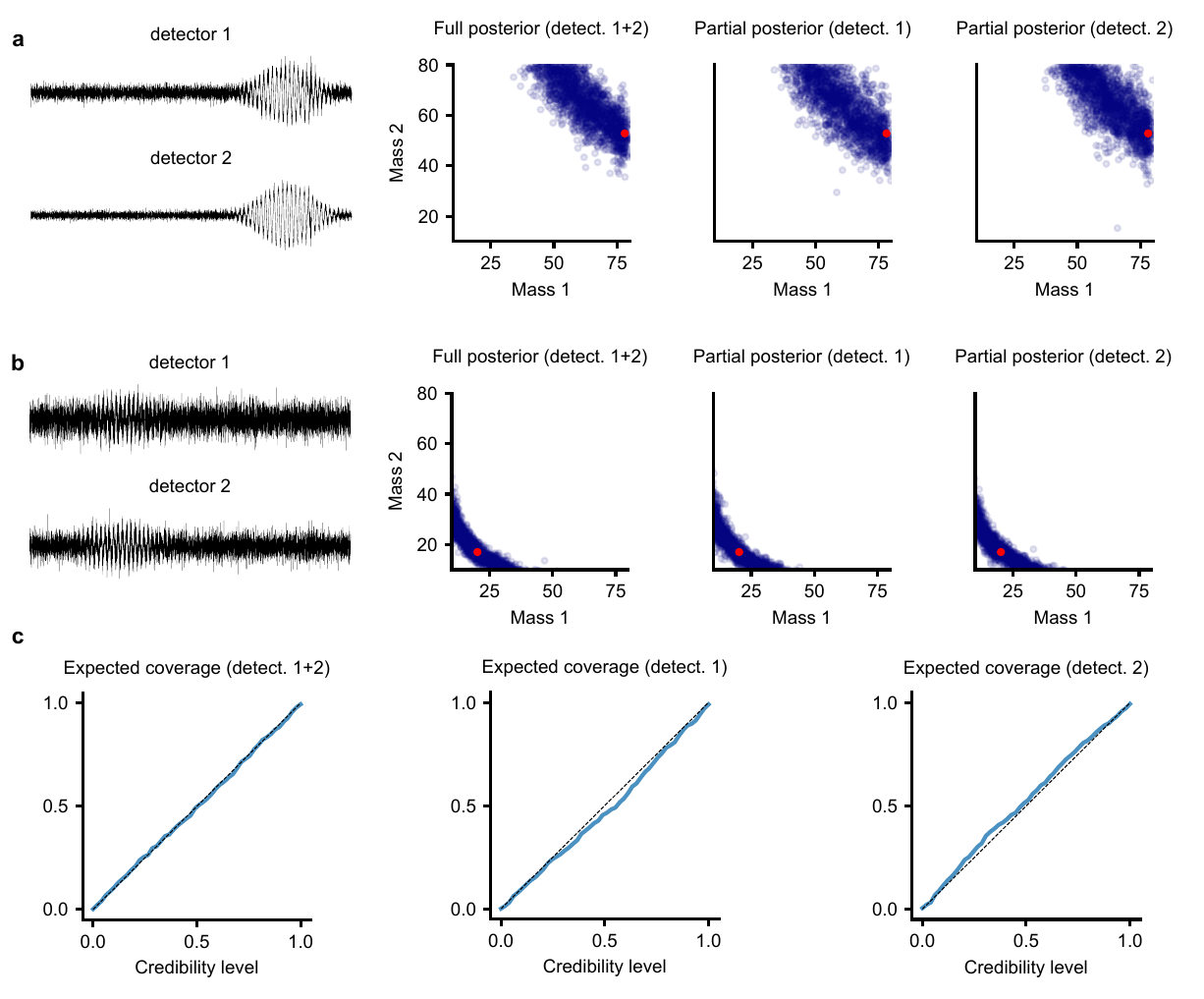}
    \caption{Inference on gravitational wave measurements. \textbf{(a)} Shows the detector measurements of a gravitational wave (first column). The associated posterior and partial posteriors for the detector measurements. \textbf{(b)} Shows the same as in (a) but for different observations. \textbf{(c)} Calibration analysis of the posterior and partial posteriors in terms of \textit{expected coverage}.}
    \label{fig:gw}
    \vspace{-1cm}
\end{figure}

As a test case, we will consider the Gravitational Waves benchmark tasks as presented in \citet{hermans2022trust}. In this case, we have low dimensional $\mathbf{\theta} \in \mathbb{R}^2$, i.e., the masses of the two black holes, and two high dimensional $\vx \in \mathbb{R}^{8192}$ measurements of the corresponding gravitational waves from two different detectors. In this case, it is clear that learning the likelihood, i.e., a conditional generative model for the high dimensional observations, is harder than just learning the posterior over the two parameters. A common practice for high dimensional observations is to use an \textit{embedding network}, i.e., a neural network that compresses it to a lower dimensional vector. \citet{hermans2022trust} did use a convolutional embedding net for NPE on this task. As already hinted in the manuscript, we can do the same for Simformer, i.e., we compress the detector measurements using a convolutional neural network into a single token. Additionally to the full posterior distribution, we are still interested in the partial posterior distributions as, e.g., there might only be measurements from one of the detectors (notably, the measurements are not independent). We hence only target the conditionals $p(\vtheta|\vx_1, \vx_2)$, $p(\vtheta|\vx_1)$ and $p(\vtheta|\vx_2)$. We use 100k simulations for training. For two examples, we show the estimated (partial) posterior(s) in Fig.~\ref{fig:gw}a Fig.~\ref{fig:gw}b. Simformer can combine the information from both detectors in a meaningful way (as verified by a calibration analysis, Fig.~\ref{fig:gw}c.

\newpage

\subsection{Details on general guidance}

Diffusion guidance can vary in its implementation from less rigorous to highly rigorous approaches. Achieving rigor in this context typically necessitates a known likelihood function. However, in the realm of SBI, this likelihood function is often either intractable or challenging to compute \citep{chung2023diffusion}. Consequently, our focus is directed towards universally applicable approximations, as discussed in the works of \citet{lugmayr2022repaint} and \citet{bansal2023universal}.

In our methodology, we integrate two principal strategies that have demonstrated efficacy in practical scenarios. The first of these strategies is self-recurrence, as advocated by \citet{lugmayr2022repaint}. This might also be interpreted as a predictor-corrector algorithm \citep{song2020score} with a pseudo-Gibbs sampling corrector. This approach has been shown to substantially improve performance, though it necessitates an increase in computational resources. The second strategy entails adjusting the estimated score with a general constraint function, which we evaluate on a \textit{denoised} estimate of the variables~\citep{bansal2023universal, chung2023diffusion, rozet2021arbitrary}. Overall, this is remarkable flexibility and supports almost any constraint to be incorporated. We provide a pseudo-code in Algorithm \ref{alg:general_guidance}. In our experimental assessments, it proved to be sufficiently accurate. For comparative purposes, we also implemented the RePaint method as proposed by \citet{lugmayr2022repaint}. However, it is important to note that this method primarily applies to normal conditioning and does not readily extend to general constraints. On the other hand, \textit{General guidance} requires the specification of a scaling function, which up and down scales the constrained score at different diffusion times $t$. As the magnitude of the marginal score does depend on the SDE, this scaling function should also. In our experiment, we generally used a scaling function of the form
$ s(t) = \frac{1}{\sigma(t)^2}$, i.e., which is inversely proportional to the variance of the approximate marginal SDE scores.

\begin{algorithm}
\caption{General Guidance}
\label{alg:general_guidance}
\begin{algorithmic}
\REQUIRE Number of steps $T$, Min time $T_{\text{min}}$, Max time $T_{\text{max}}$,self-recurrence steps $r$, scaling function $s(t)$ and constraint function $c(x)$, drift coefficient $f(x,t)$, diffusion coefficient $g(t)$, associated mean and standard deviation functions $\mu,\sigma$ such that $\vxhat_t = \mu(t)\vxhat_0 + \sigma(t)\epsilon$ .
\STATE Set time step $\Delta t = \frac{T_{\text{max}} - T_{\text{min}}}{T}$
\STATE Sample $\hat{\bm{x}}_T \sim \mathcal{N}(\mu_T, \sigma_T \mathbf{I})$ \qquad \textit{// Initialize at terminal distribution}
\FOR{$i = 1$ \textbf{downto} $T$}
    \STATE $t_i = T_{\text{max}} - i \cdot \Delta t$
    \FOR{$j = 1$ \textbf{to} $r$}
        \STATE $\epsilon \sim \mathcal{N}(0, \mathbf{I})$
        \STATE $s = s_{\phi}(\vxhat_{t_{i+1}}, t_i)$  \qquad \textit{// Marginal score estimate}
        \STATE $\vxhat_{\sim 0} = \frac{\vxhat_{t_{i+1}} + \sigma(t_{i+1})^2 \cdot s}{\mu(t_{i+1})}$ \qquad \textit{// Denoise}
        \STATE $\tilde{s} = s + \nabla_\vxhat \log \sigma (s(t)c(\vxhat_{\sim0}))$  \qquad \textit{// Constraint score}
        \STATE $\vxhat_{t_i} = \vxhat_{t_{i+1}} - \left(f(\vxhat_{t_{i+1}}, t_i) - g(t_i)^2 \cdot \tilde{s}\right) \Delta t - g(t_i) \sqrt{\Delta t} \cdot \epsilon$
        \IF{$r > 0$}  
            \STATE \textit{// Resample future point using SDE equations}
            \STATE $\epsilon \sim \mathcal{N}(0, \mathbf{I})$
            \STATE $\vxhat_{t_{i+1}} = \vxhat_{t_i} + f(\vxhat_{t_{i+1}}, t_i) \Delta t + g(t_i) \sqrt{\Delta t} \cdot \epsilon$
        \ENDIF
    \ENDFOR
\ENDFOR
\STATE \textbf{return} $\vxhat_{T_{min}}$
\end{algorithmic}
\end{algorithm}

\begin{figure}[H]
    \centering
    \includegraphics[width=\textwidth]{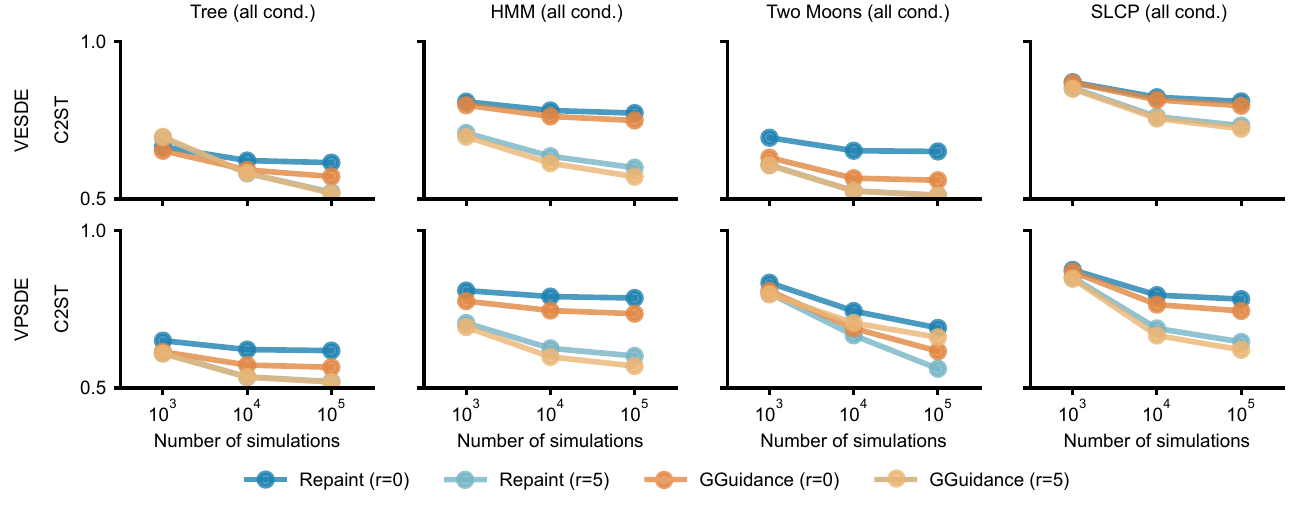}
    \caption{The Simformer exclusively trained for joint distribution estimation (i.e., $M_C$ is always zero and thereby disables model-based conditioning). As model-based conditioning is not feasible, conditioning is implemented through diffusion guidance. This figure demonstrates the application of varying levels of self-recurrence, denoted as $r$, to enforce different conditions.}

    \label{fig:guidance_bm}
\end{figure}

\paragraph{Benchmarking the Guidance Methods:} In this experiment, we diverged from traditional approaches by training the Simformer exclusively for joint estimation. The primary distinction from a conditional distribution lies in the condition mask distribution, which in this case is a point mass centered at the all-zero vector. Our comparative analysis, as depicted in Figure~\ref{fig:guidance_bm}, reveals that diffusion guidance-based methods fall short in performance when operating within the same computational budget and without self-recurrence. A notable observation is that the application of self-recurrence markedly improves the results, aligning them closely with those achieved through model-based conditioning. This enhancement, however, incurs a fivefold increase in computational demand.

\begin{figure}[H]
    \centering
    \includegraphics[width=\textwidth]{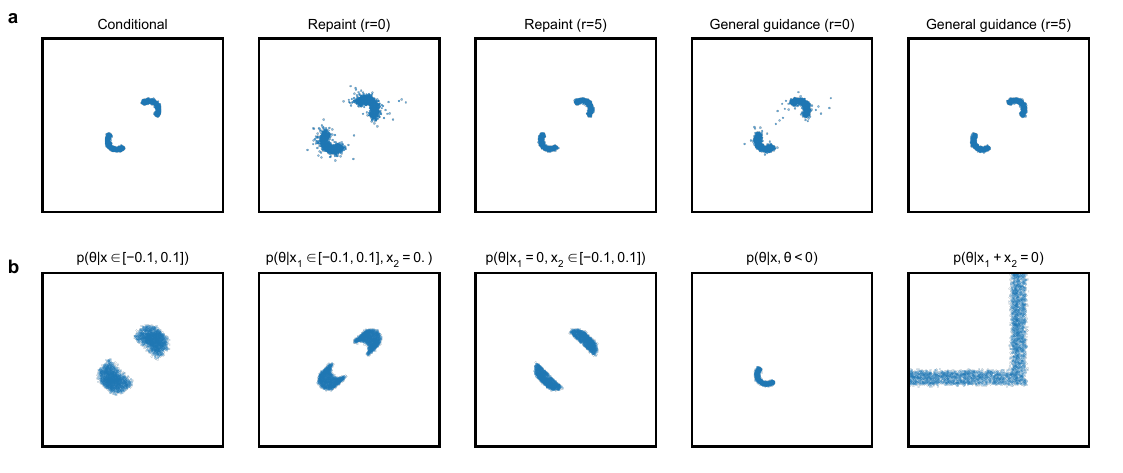}
    \caption{\textbf{(a)} Shortcomings of diffusion guidance without self recurrence $r=0$, which can be fixed using $r=5$. This, however, also increases the computational cost by five. \textbf{(b)} General set constraints enforced using diffusion guidance for the Two Moons tasks. The (conditional) Simformer model was trained on $10^5$ simulations. Any exact condition was model-based, and any set constraint was enforced through guidance.}
    \label{fig:guidance_two_moons}
\end{figure}

\paragraph{Arbitrary Constraints:}
The above benchmarks have demonstrated the high accuracy potential of diffusion guidance. The effectiveness of diffusion guidance in accurately reconstructing distributions is evident from Figure~\ref{fig:guidance_two_moons}a. Despite its general efficacy, the model exhibits minor issues, such as the slightly excessive noise observed in the two-moon scenario. These issues, however, can be mitigated through the application of self-recurrence. Figure~\ref{fig:guidance_two_moons}b further illustrates our approach's capability to concurrently address multiple constraints while also being able to integrate model-based conditioning (every exact constrained is model-based).

\label{sec:appendix_guidance}

\end{document}